\PassOptionsToPackage{table}{xcolor}
\documentclass{article}
\usepackage[T1]{fontenc}
\usepackage{iclr2027_conference,times}

\usepackage{amsmath,amsfonts,bm}

\def\eqref#1{equation~\ref{#1}}
\def\1{\bm{1}}

\DeclareMathAlphabet{\mathsfit}{\encodingdefault}{\sfdefault}{m}{sl}
\SetMathAlphabet{\mathsfit}{bold}{\encodingdefault}{\sfdefault}{bx}{n}

\usepackage{amsmath}
\usepackage{amssymb}
\usepackage{booktabs}
\usepackage{graphicx}
\usepackage{longtable}
\usepackage{multirow}
\usepackage{xcolor}
\usepackage{hyperref}
\usepackage{url}
\usepackage{etoolbox}
\definecolor{claudecodebg}{RGB}{253,244,233}
\definecolor{codexbg}{RGB}{235,247,239}
\definecolor{openclawbg}{RGB}{235,244,252}
\newcommand{\cmark}{\ensuremath{\checkmark}}

\newcommand{\NumTasks}{50}
\newcommand{\NumCandidates}{82}
\newcommand{\NumScenarios}{19}
\newcommand{\NumFacts}{3{,}102}
\newcommand{\NumAtoms}{6{,}414}
\newcommand{\VideoHours}{17.1}

\newcommand{\CoverageSaturation}{0.18}
\newcommand{\DirectionCountCap}{15}
\newcommand{\ContentWeight}{0.50}
\newcommand{\CitationWeight}{0.30}
\newcommand{\PresentationWeight}{0.20}
\newcommand{\VideoQueryWeights}{(0.35,0.35,0.20,0.10)}
\newcommand{\UserQueryWeights}{(0.30,0.30,0.20,0.20)}
\newcommand{\VideoWindowSeconds}{300}
\newcommand{\MinExternalSources}{25}
\newcommand{\MinRelatedVideos}{5}
\newcommand{\MinDocuments}{3}
\newcommand{\MinSourcesPerAnchor}{3}
\newcommand{\MaxRepairRounds}{3}
\newcommand{\JudgeVotes}{3}
\newcommand{\JudgeTemperature}{0}
\newcommand{\CalibrationGap}{20\%}

\newcommand{\NumSystems}{18}
\newcommand{\BestAllTaskScore}{63.3}
\newcommand{\MeanContentScore}{45.4}
\newcommand{\MeanCitationScore}{47.8}
\newcommand{\MeanPresentationScore}{81.8}
\newcommand{\MeanFactsPerTask}{62}
\newcommand{\MeanAtomsPerTask}{128}
\newcommand{\MeanReportHours}{2}
\newcommand{\SweepAgentHours}{100}
\newcommand{\NumReportAttempts}{900}

\newcommand{\HumanRoundOneAccuracy}{64.0\%}
\newcommand{\HumanRoundOneUnanimity}{77.0\%}
\newcommand{\HumanRoundTwoAccuracy}{87.0\%}
\newcommand{\HumanRoundTwoUnanimity}{81.0\%}
\newcommand{\MaxRequestRetries}{5}
\newcommand{\MaxReportAttempts}{3}
\newcommand{\ReportTimeoutHours}{10}

\title{Beyond Watching and Searching: Benchmarking Agents for Video-Centered Deep Research Reports}

\newcommand{\OmniDRAuthors}{%
  \textbf{Jiahao Meng\textsuperscript{1,2}\textsuperscript{\textdagger}, Qize Yang\textsuperscript{2},
  Yuxuan Wang\textsuperscript{2}, Yunfei Chu\textsuperscript{2}}, 
  \textbf{Wensi Dai\textsuperscript{3}, Jin Xu\textsuperscript{2,*},
  Yunhai Tong\textsuperscript{1,*}}\\
  {\normalfont\small\textsuperscript{1}Peking University\quad
  \textsuperscript{2}Alibaba Token Hub, Alibaba Group\quad \textsuperscript{3}Independent Researcher}\\
  {\normalfont\small
  \textsuperscript{*}Co-corresponding authors}%
}
\newcommand{\OmniDRReleaseStatement}{The code and data are released at \url{https://github.com/marinero4972/OmniDR-Bench}.}
\author{\OmniDRAuthors}
\makeatletter
\patchcmd{\@maketitle}
  {Anonymous authors\\Paper under double-blind review}
  {\@author}
  {}
  {\PackageError{OmniDR}{Could not replace the author block}{Check the ICLR template.}}
\makeatother

\iclrfinalcopy

\begin{document}

\maketitle
% First-author internship note; dagger avoids clashing with the * corresponding mark.
\begingroup
\renewcommand{\thefootnote}{\textdagger}
\footnotetext{Work done as an intern at Qwen Team, Alibaba Token Hub.}
\endgroup
% Running head for the arXiv release.
\fancyhead[L]{Preprint}
\fancyhead[C]{}
\fancyhead[R]{}
\fancyfoot[C]{\thepage}
\pagestyle{fancy}
\thispagestyle{fancy}

\begin{abstract}
Deep research agents synthesize open-web evidence into long reports, but current benchmarks do not evaluate research centered on a source video.
We introduce \textbf{Video-Centered Deep Research Report Generation}.
Given a video and user request, an agent identifies research questions, retrieves external multimodal evidence, and produces an interactive report linked to relevant video moments.
This task has direct practical value because it helps users locate relevant moments, verify claims, reproduce demonstrated procedures, and make informed decisions from video content.
We present a skill-based workflow and \textbf{OmniDR-Bench} for this task.
For evaluation, we propose \textbf{Omni Atomic Rubric Trees} (Omni-ARTs).
Each tree contains three dimensions, 23 shared directions, and task-specific atomic binary criteria.
The directions test audiovisual grounding, multimodal evidence use, and interactive presentation.
\textbf{Audited Rubric Construction} (ARC) builds these trees through independent evidence collection, automated checks, behavioral calibration, and iterative human review.
OmniDR-Bench contains \NumTasks{} Chinese and English tasks across \NumScenarios{} real-world subcategories.
It includes \NumAtoms{} criteria.
Across \NumSystems{} agent configurations, the best system scores \BestAllTaskScore{} out of 100.
Current agents still need broader content coverage, stronger causal and logical analysis across modalities, and more accurate citation support.
We further analyze models, harnesses, skill configurations, video categories, and research trajectories.
\ifdefined\OmniDRReleaseStatement
\OmniDRReleaseStatement
\fi

\end{abstract}

\section{Introduction}
\label{sec:introduction}

Videos teach knowledge, share opinions, demonstrate procedures, and advertise products through speech, imagery, and temporal context.
Compared with text, video natively preserves richer situational context, visible entities, spatial relations, and fine-grained actions that are difficult to verbalize completely.
These cues can reveal questions about mechanisms, credibility, alternatives, and practical use that a broad text topic may not expose.
Viewers may therefore need to locate relevant moments, verify claims and background, discover related material, or apply the content to a real decision.
These needs call for research that remains centered on the source video.

\begin{figure*}[t]
  \centering
  \includegraphics[width=0.99\textwidth]{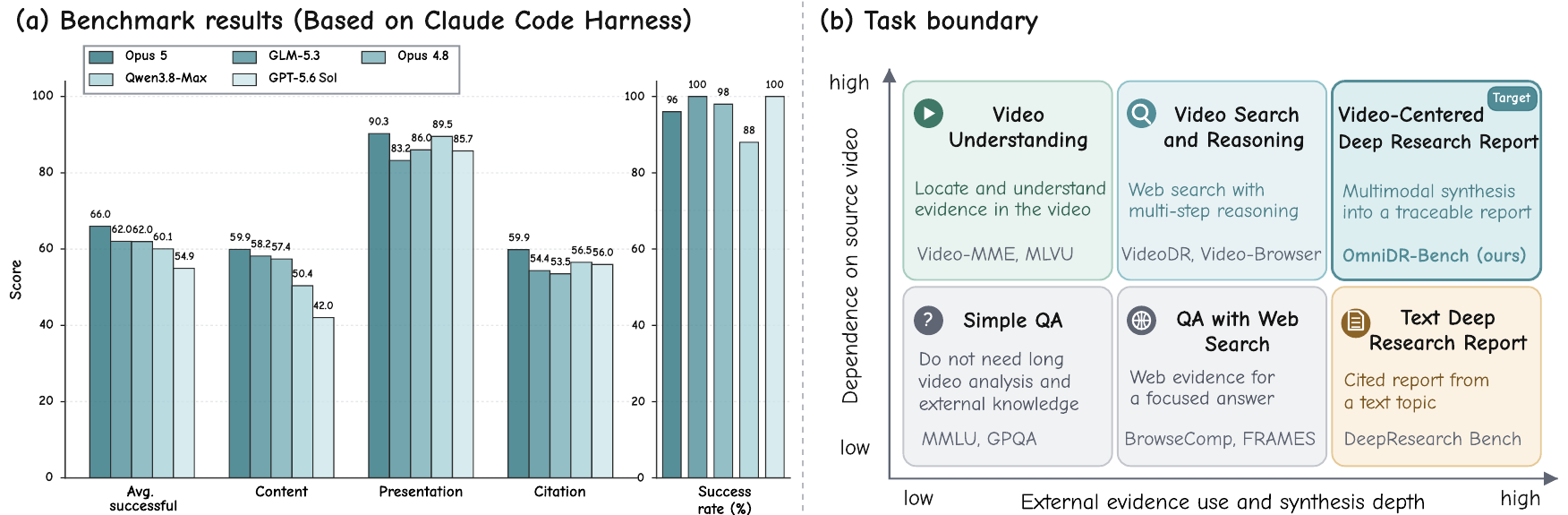}
  \caption{Video-Centered Deep Research and representative OmniDR-Bench results.
  Left: successful-report scores and task success.
  Right: the task integrates source-video audiovisual context with external text, image, and video evidence in an interactive, cited report.}
  \label{fig:teaser}
\end{figure*}

Existing agents address only parts of this goal.
Text-centered deep research agents produce cited reports through iterative web search and synthesis \citep{java2025characterizing,li2025websailor,qiao2025webresearcher}.
They do not require the audiovisual or temporal evidence of a given video.
Video deep research combines source-video evidence with open-web retrieval and multi-step reasoning \citep{liang2025videobrowser,liu2026videodr,gao2026videosearcher,fang2026videodeepresearch,liu2026videorover,zhang2026adavdr}.
Its benchmarks mainly target difficult question answering (QA) tasks with short answers.
Text-image research report systems \citep{shi2026vidr,zhang2026ptah,ma2026tvir} synthesize multimodal web material without taking a source video or retrieved videos as research evidence.
Figure~\ref{fig:teaser} locates these settings by input, research depth, and output form.
Table~\ref{tab:task_comparison} further compares their operational boundaries.
No existing setting evaluates a complete research report that is organized around one source video and enriched by external multimodal evidence.

We introduce \textbf{Video-Centered Deep Research Report Generation} to fill this gap.
Given a source video and a user request, an agent identifies research-worthy claims, procedures, parameters, entities, and open questions.
It investigates webpages, documents, images, and related videos, then connects this evidence to relevant source moments.
The output is an interactive HTML report for understanding, verification, and practical application.
We implement a skill-based reference workflow with specialized media readers, anchor-conditioned research, evidence verification, and interactive rendering.

Evaluating this task is difficult because an open-ended report has no unique reference answer.
Valid reports may use different organizations and source sets \citep{du2025deepresearchbench,sharma2025researchrubrics,zhong2026draco}.
An evaluator must still distinguish citation volume from claim support and assess video grounding, cross-source reasoning, and the rendered artifact \citep{zhao2026reportlogic,leung2026citationverifier,an2026aeseval}.
We address this challenge with \textbf{Omni Atomic Rubric Trees} (Omni-ARTs).
Each tree has three levels beneath the aggregate score: three quality dimensions, 23 shared diagnostic directions, and task-specific atomic binary criteria.
The hierarchy preserves comparable upper-level scores while testing the evidence needs of each task.
Reliable trees also require a sufficiently broad and independently verified evidence base.
We therefore propose \textbf{Audited Rubric Construction} (ARC), which collects multimodal evidence, screens candidate criteria, calibrates score behavior, and verifies every fact and criterion through iterative human review.
Together, Omni-ARTs and ARC define \textbf{OmniDR-Bench}.
Here, \emph{Omni} denotes research across video, audio, image, and text, while \emph{DR} denotes Deep Research.

OmniDR-Bench contains \NumTasks{} tasks across \NumScenarios{} subcategories and \NumAtoms{} criteria.
Experiments compare \NumSystems{} configurations that pair models with agent harnesses, the execution frameworks that manage planning, context, delegation, and recovery.
The strongest system achieves \BestAllTaskScore{} on the $0$--$100$ rubric scale.
Presentation averages \MeanPresentationScore{} points, while content and citation average only \MeanContentScore{} and \MeanCitationScore{} points.
Under the shared report template, current agents satisfy presentation requirements more consistently than audited coverage, reasoning, and citation-support criteria.

Our contributions are threefold.
\textbf{First}, we propose Video-Centered Deep Research Report Generation, extending video deep research from short answers to complete and traceable reports.
\textbf{Second}, we propose Omni-ARTs and ARC for fine-grained report evaluation and reliable benchmark construction.
\textbf{Third}, we release OmniDR-Bench and a reference agent workflow.
We analyze models, agent harnesses, skill configurations, video categories, and long-horizon research trajectories.

\section{Video-Centered Deep Research}
\label{sec:task}

This section defines the input, output, and evidence requirements of Video-Centered Deep Research Report Generation.
It then presents a skill-based agent workflow that implements the full research process.

\subsection{Task Definition}
\label{sec:task_definition}

An instance contains a source video $V$, a user request $Q$, and an open tool environment $\mathcal{E}$.
An agent $\mathcal{A}$ produces an interactive HTML report
\begin{equation}
H=\mathcal{A}(V,Q,\mathcal{E}),
\end{equation}
where $H$ is the delivered report.
This task helps users locate important material, verify claims and context, and apply video content to practical decisions.
These uses require more than a recap.
The report must preserve what the video shows and says while adding external research that addresses the user's need.

\begin{figure*}[t]
  \centering
  \includegraphics[width=\textwidth]{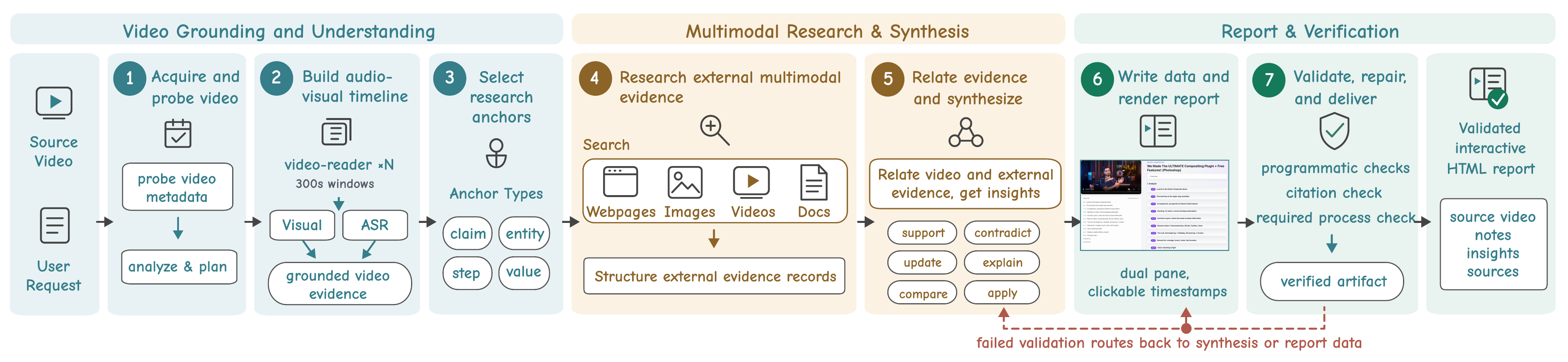}
  \caption{Seven-stage reference workflow.
  Research anchors connect source-video moments to external webpage, image, video, and document evidence.
  The agent synthesizes these records into an interactive report and repairs failed checks.}
  \label{fig:agent_workflow}
\end{figure*}

Figure~\ref{fig:agent_workflow} makes this process concrete.
Stage 3 turns request-relevant claims, procedures, parameters, entities, and open questions into research anchors.
Stage 4 uses these anchors to retrieve evidence from webpages, images, videos, and documents.
Stage 5 records how the external evidence supports, contradicts, updates, explains, compares, or applies the video content.

A valid report keeps the given video central and contributes research beyond a summary.
Important findings must connect to both external sources and relevant video moments.
Its text, media, citations, and interactions must let readers inspect the evidence.
Together, these requirements capture video centrality, research depth, multimodal evidence use, traceability, and usability.
We accept a task only when video-only analysis is insufficient and generic web research cannot replace the source video.
Sufficient public evidence must also exist within a practical budget.
Table~\ref{tab:task_comparison} distinguishes our task from text-image report systems.
Those systems start from a textual topic and synthesize webpages, images, and figures without reading a source video or retrieving external video evidence.

\begin{table*}[t]
\caption{Task boundaries.
Ours uses one given video as organizing evidence for an auditable report.}
\label{tab:task_comparison}
\centering
\small
\setlength{\tabcolsep}{6.0pt}
\renewcommand{\arraystretch}{1.08}
\resizebox{0.98\textwidth}{!}{%
\begin{tabular}{@{}lcccccc@{}}
\toprule
Setting & Source video & Open web & Video retrieval & Long report & Time links & Primary objective \\
\midrule
Video summary & Yes & No & No & No & No & Condense one video \\
Text deep research & No & Yes & No & Yes & No & Research a standalone topic \\
Open-web video search & No & Yes & Yes & No & No & Find video evidence \\
Video deep research QA & Yes & Yes & No & No & No & Answer one hard question \\
Text-image research report & No & Yes & No & Yes & No & Synthesize text and image evidence \\
\textbf{Ours} & \textbf{Yes} & \textbf{Yes} & \textbf{Yes} & \textbf{Yes} & \textbf{Yes} & \textbf{Research and audit one video} \\
\bottomrule
\end{tabular}
}
\end{table*}

\subsection{Agent Workflow}
\label{sec:workflow}

We package the seven-stage agentic workflow as a high-level \emph{video-research skill}, a reusable instruction and tool module.
It belongs to a suite of thirteen skills and invokes specialized \texttt{video-reader} and \texttt{doc-reader} modules as subagents.
The main agent carries the timeline, research anchors, external evidence, cross-source relations, unresolved questions, and report data across stages.
This shared state reduces repeated context loading and helps prevent goal drift in long trajectories \citep{wang2026trajectoryerrors}.

Figure~\ref{fig:agent_workflow} summarizes the seven stages from video intake through report validation and repair.
The agent first acquires the video and aligns visual observations with speech on one timeline.
It then selects request-relevant anchors with timestamps or frame evidence.
Parallel branches inspect webpages, images, videos, and documents.
Each branch returns structured evidence with provenance.
The main agent connects this evidence to the source video and synthesizes supported findings.
It writes structured report data and renders an interactive report with video-time navigation.
The final stage runs programmatic, citation, and required-process checks.
Failed checks return to evidence synthesis or report data for targeted repair.
Appendix~\ref{app:workflow} expands this control-flow overview with the reader roles, state fields, browser checks, evidence verification, and released budgets.
Appendix Figure~\ref{fig:qualitative_report} shows the final interactive artifact for one complete example.

\section{OmniDR-Bench}
\label{sec:benchmark}

OmniDR-Bench evaluates open-ended reports with Omni Atomic Rubric Trees, separating video-grounded content, citation support, and interactive presentation.
Audited Rubric Construction builds and verifies the evidence and criteria for each tree.

\subsection{Video Collection and Query Construction}
\label{sec:data_collection}

We select four families that reflect common reasons for researching a video.
Knowledge and viewpoint videos require explanation and verification.
Consumer and business videos support comparison and decision making.
Physical procedure videos require faithful reconstruction and diagnosis.
Creative and digital videos emphasize workflows, tools, and transferable practice.
Together, these families test distinct evidence needs and report outcomes.
We manually source and download candidate videos from YouTube and Bilibili.
Each retained video contains claims, steps, parameters, entities, or open questions that depend on visual or temporal context.
The fixed benchmark contains \NumTasks{} Chinese and English tasks across \NumScenarios{} subcategories, with the full collection and curation protocol documented in Appendix~\ref{app:construction}.

\begin{figure*}[t]
  \centering
  \includegraphics[width=0.98\textwidth]{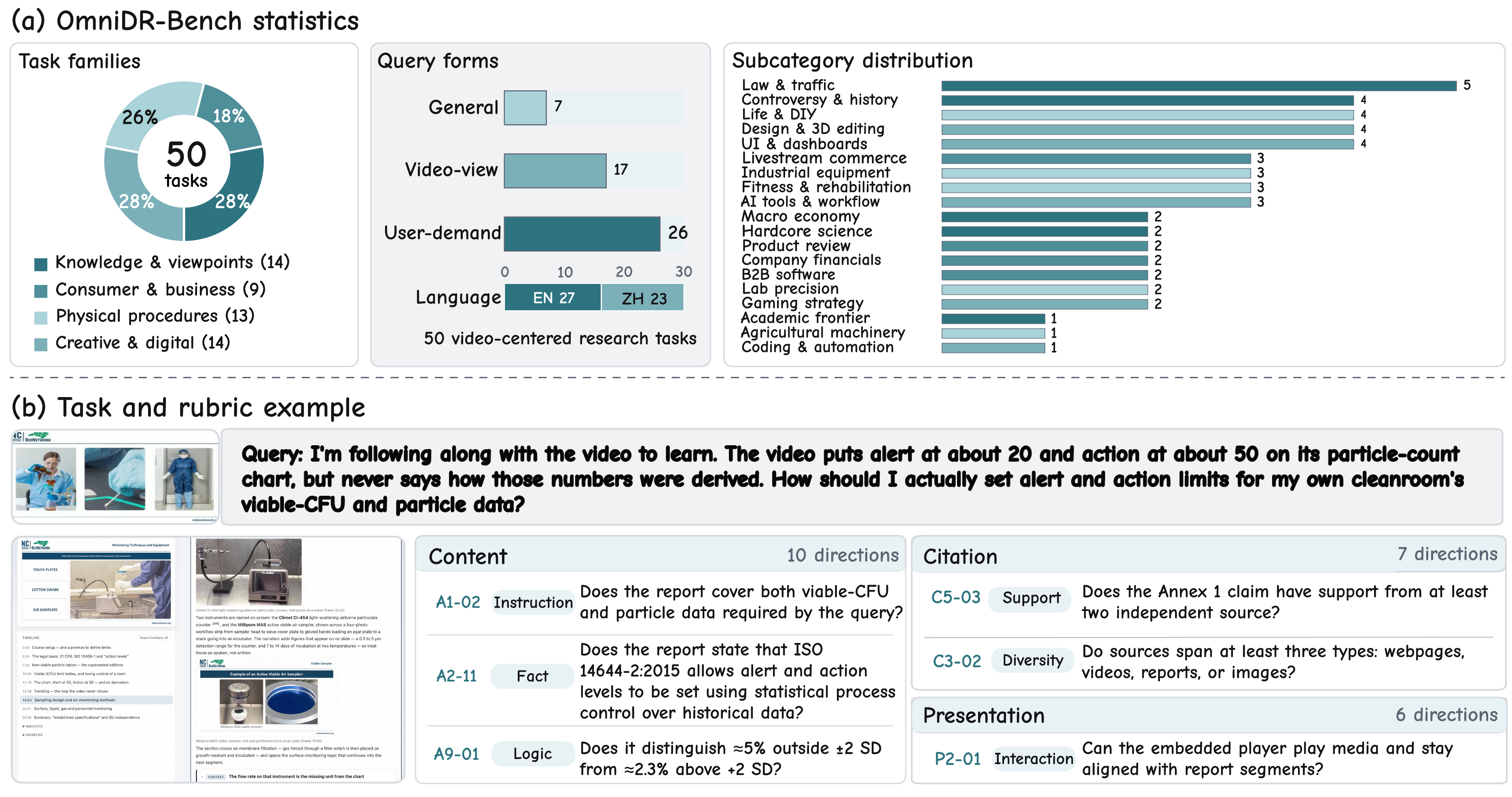}
  \caption{OmniDR-Bench statistics and one task that connects source-video evidence to external multimodal research.
  Selected content, citation, and presentation criteria show the evaluation scope.
  Appendix Tables~\ref{tab:query_examples} and~\ref{tab:qualitative_tasks} give more tasks, while Appendix Figures~\ref{fig:rubric_tree} and~\ref{fig:qualitative_report} show the full Omni-ART hierarchy and one complete report.}
  \label{fig:benchmark_overview}
\end{figure*}

The query pool combines a general report request with two focused views.
The \emph{video view} proposes research directions around theories, cases, objects, or claims.
The \emph{user demand view} formulates a concrete decision, diagnosis, reproduction, or application need.
Annotators rewrite selected candidates and require each query to need both the given video and external evidence while remaining answerable from public sources.
Figure~\ref{fig:benchmark_overview} summarizes the resulting coverage, and Appendix~\ref{app:queries} gives the query scoring, acceptance tests, and examples.

Each task averages \MeanAtomsPerTask{} criteria, giving \NumAtoms{} fine-grained checks.
One report takes roughly \MeanReportHours{} hours, so one system requires about \SweepAgentHours{} agent-hours before repeated grading.
The deliberate 50-task scale covers four families and 19 subcategories while keeping the \NumSystems{}-configuration comparison tractable and favoring dense diagnosis over short-answer volume.
Appendix~\ref{app:taxonomy} lists all 19 subcategories, their definitions, and associated research needs.

\subsection{Omni Atomic Rubric Trees}
\label{sec:benchmark_construction}

An \textbf{Omni Atomic Rubric Tree} (Omni-ART) has three rubric levels beneath the final aggregate score.
The first level contains content, citation, and presentation.
The second level contains 23 shared diagnostic directions.
The third level contains task-specific criteria under each applicable direction.
The \emph{Omni} scope covers video, audio, image, and text evidence and the relations among them.
Its directions therefore assess audiovisual faithfulness, cross-modal reasoning, multimodal source use, claim support, image-text correspondence, and video-time interaction.
This hierarchy supports comparison at shared upper levels without prescribing one report structure or source set.
Figure~\ref{fig:rubric_tree} in Appendix~\ref{app:evaluation} shows the complete hierarchy, gate assignments, and model or deterministic evaluator types.
Each leaf is an atomic binary criterion with an expected outcome, evaluator, and evidence difficulty.
Semantic leaves bind to independent fact records; structural leaves bind to executable program or browser checks, making each decision auditable.

We propose \textbf{Audited Rubric Construction} (ARC) to build and verify each Omni-ART.
Evaluated reports never enter ARC evidence acquisition or criterion generation.
This separation matters because evaluated reports can omit valid facts or favor the sources selected by one system.
ARC therefore builds an independent fact library before writing any criterion.
For each task, isolated agents first produce a Video Evidence Map of timestamped audiovisual facts.
They independently search webpages, documents, images, and related videos to build a Web Evidence Map.
This external multimodal search adds visual analogues, demonstrations, primary records, standards, and updated context that text search alone may miss.
ARC then connects the two maps through support, contradiction, update, explanation, comparison, and application relations.
The resulting fact library makes cross-modal evidence explicit before any rubric leaf is written.
Omni-ART scores the submitted report rather than compliance with the reference workflow in Section~\ref{sec:workflow}.
Any internal process can receive credit when its final artifact satisfies the same criteria.
The fact library specifies salient research targets; it does not prescribe the report's organization or an identical set of retrieved sources.

\begin{figure*}[t]
  \centering
  \includegraphics[width=0.97\textwidth]{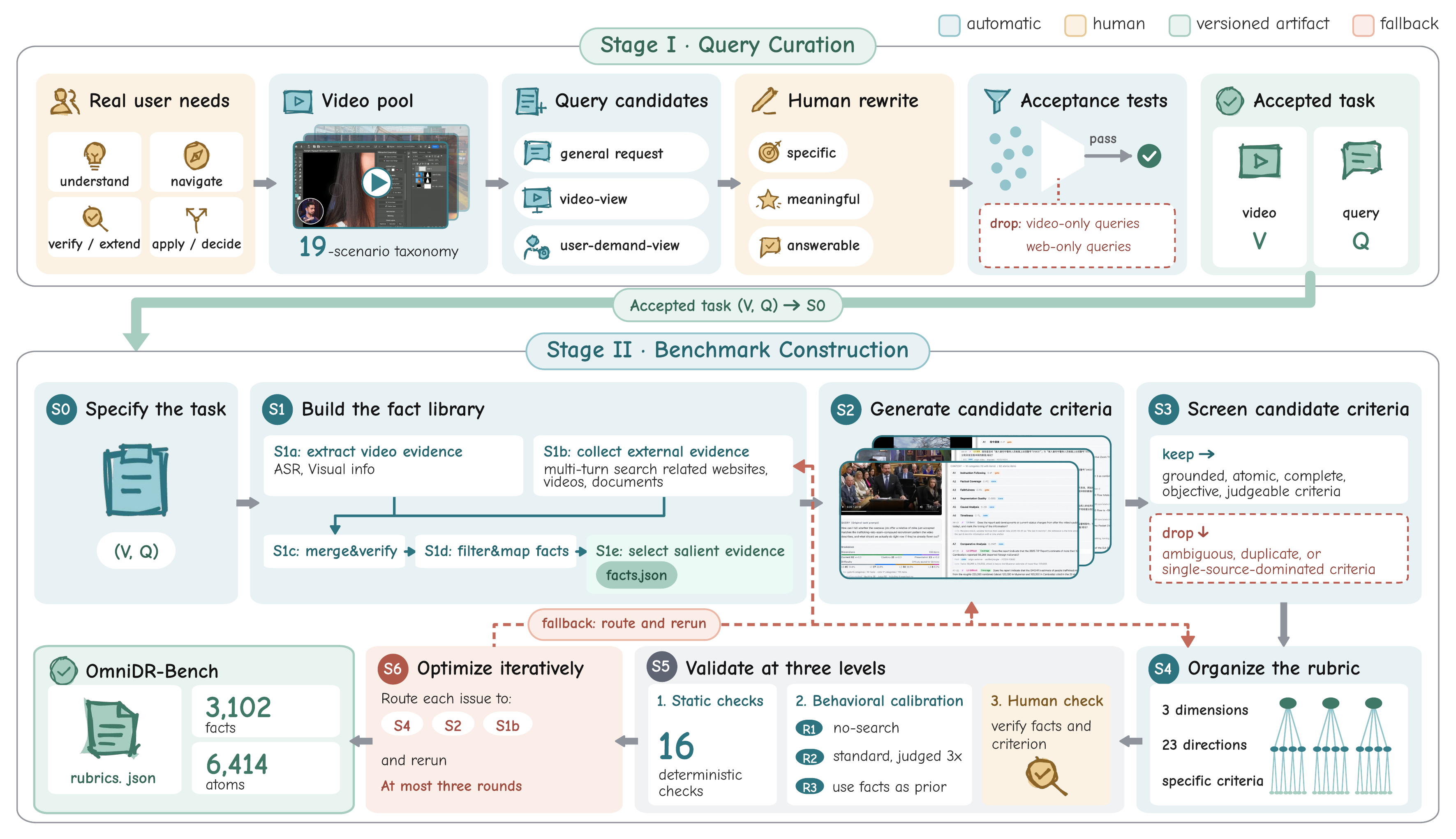}
  \caption{Query curation and Audited Rubric Construction (ARC).
  Curation retains source-video-dependent tasks that require external evidence.
  ARC independently gathers audiovisual and open-web multimodal evidence, derives an Omni Atomic Rubric Tree (Omni-ART), and validates both facts and criteria.}
  \label{fig:benchmark_construction}
\end{figure*}

Figure~\ref{fig:benchmark_construction} summarizes the core ARC path from task specification to a versioned benchmark artifact.
S0 fixes the task scope and search routes to prevent evidence collection from drifting toward the broad topic.
S1 collects video evidence and external multimodal evidence in isolated branches to preserve provenance and keep either branch from copying the other's framing.
It then reconciles support and conflict, removes irrelevant or redundant records, and retains salient evidence for the user request.
S2 converts this fact library and the report requirements into binary candidate criteria so that each leaf can be checked directly.
Generated criteria can be ambiguous, duplicated, weakly supported, or concentrated in a few directions.
S3 retains criteria that are grounded in evidence or executable behavior, atomic, complete enough to decide without unstated context, objectively worded, and judgeable by the assigned evaluator.
It also checks uniqueness, source support, and direction coverage.
S4 assembles and caps the retained leaves under the three dimensions and 23 directions so that high-volume directions do not dominate the score.

S5 combines static checks, behavioral calibration, and human artifact verification.
Static checks remove schema, identifier, evidence-binding, and logic errors before any behavioral test.
Behavioral calibration compares a video-only report $R_1$, a full-tool research report $R_2$, and a fact-library-assisted report $R_3$.
All three reports are generated by Claude Opus 4.8 under Claude Code.
$R_2$ uses the full tool set and video-research skill.
Every retained task satisfies $R_1<R_2<R_3$ and $R_3-R_1>\delta_{\mathrm{cal}}$.
Here, $\delta_{\mathrm{cal}}$ is the minimum required gain from $R_1$ to $R_3$.
These three report levels check whether the Omni-ART responds to additional evidence access.
We also score each $R_2$ report three times.
Appendix Table~\ref{tab:construction_artifacts} summarizes this control with the other S5 checks.
Only $0.89\%$ of applicable criteria receive non-unanimous decisions on average, which indicates stable evaluation.
Human reviewers inspect the facts and criteria and return semantic errors to S6.
S6 routes repairable issues back to their originating stage, removes candidate packages with pervasive artifact errors, and advances accepted packages to the next validation round.
After at most three rounds, ARC freezes the evidence, Omni-ART, and evaluator versions used in the released benchmark.
The released snapshot contains \NumAtoms{} criteria.
Appendix~\ref{app:construction} details the persistent artifacts, isolated evidence collection, provenance and conflict handling, pruning rules, screening logic, calibration, and repair process.
Appendix~\ref{app:human_protocol} reports the human verification protocol and agreement.
\looseness=-1

\subsection{Evaluation Methodology}
\label{sec:evaluation}

Table~\ref{tab:evaluation_dimensions} defines the dimension and direction layers of an Omni-ART.

\begin{table*}[t]
\caption{Evaluation dimensions.
Each direction contains task-specific binary criteria.}
\label{tab:evaluation_dimensions}
\centering
\small
\setlength{\tabcolsep}{4pt}
\renewcommand{\arraystretch}{0.94}
\resizebox{0.97\textwidth}{!}{%
\begin{tabular}{@{}p{0.125\textwidth}p{0.91\textwidth}@{}}
\toprule
Dimension & Directions \\
\midrule
Content & Instruction following, fact coverage, video faithfulness, segmentation, causal analysis, timeliness, comparison, cross-source linkage, logical consistency, ethics and safety \\
Citation & Source count, citation closure between inline citations and source records, source diversity, source authority, claim support, source freshness, multimodal material \\
Presentation & Video-note alignment, interactivity, visual quality, image-text match, readability, format use \\
\bottomrule
\end{tabular}
}
\end{table*}

The evaluator uses two check types.
J denotes semantic and visual decisions from a multimodal model judge.
M denotes deterministic programs, including browser probes, that inspect files, structure, sources, assets, navigation, citation jumps, playback, and controls.
Appendix~\ref{app:evaluation} defines every direction and evaluator contract.

For direction $d$, let $A_d$ be its applicable leaves after the guarded N/A procedure below, and let $a_j\in\{0,1\}$ be the outcome of leaf $j$.
Its raw and credited rates are
\begin{equation}
\label{eq:direction_rates}
r_d=\frac{1}{|A_d|}\sum_{j\in A_d}a_j,
\qquad
c_d=\begin{cases}
\min(1,r_d/\tau_d), & d\in\mathcal{H},\\
r_d, & \text{otherwise},
\end{cases}
\end{equation}
where $\mathcal{H}$ contains fact coverage, causal analysis, and comparison, and $\tau_d$ is the pass-rate threshold for full credit.
Let $g_d$ be zero when direction $d$ is a gate and fails, and one otherwise.
Let $w_d$ be the number of applicable leaves, capped by $\kappa_d$ for directions in $\mathcal{H}$.
For dimension $k\in\{C,R,P\}$ with direction set $\mathcal{D}_k$, the dimension and report scores are
\begin{equation}
\label{eq:score_aggregation}
D_k=\frac{\sum_{d\in\mathcal{D}_k}w_dg_dc_d}{\sum_{d\in\mathcal{D}_k}w_d},
\qquad
S=100\sum_{k\in\{C,R,P\}}\lambda_kD_k.
\end{equation}
Here, $C$, $R$, and $P$ index content, citation, and presentation, respectively.
The nonnegative dimension weights $\lambda_k$ sum to one.
Appendix~\ref{app:hyperparameters} reports all released values.
Appendix~\ref{app:score_robustness} reports task-bootstrap uncertainty and sensitivity to the released values of $\lambda_k$, $\tau_d$, and $\kappa_d$.

For an omitted fact, failed citation criteria become N/A only if a direction-level guard finds that the content gain from stating it exceeds the citation gain from removing failed criteria; otherwise, they retain zero scores.
Appendix~\ref{app:na_guard} details rollback and saturation handling.
Evaluator errors are excluded from the corresponding denominator.
A missing or unreadable report receives zero in the all-task average.

\section{Experiments}
\label{sec:experiments}

We compare current agents, their failure modes, and their behavior across models and harnesses.

\subsection{Experimental Setup}
\label{sec:experimental_setup}

\noindent\textbf{Implementation details.}
Each run receives the same video, query, tools, and report contract, with a shared interface template specifying the required HTML structure and behavior.
Benchmark evidence remains hidden.
Agents may inspect the video, search multimodal sources, and test the report in a browser.
GPT-5.6 Sol judges semantic and visual criteria because it supports multimodal inputs, bilingual instructions, and structured outputs.
Deterministic checks, including browser probes, verify observable properties.
A run succeeds when its final report is recoverable and evaluable.
Across the \NumReportAttempts{} model--task attempts, 835 reports are evaluable and 65 are unsuccessful or unavailable; 19 of the latter are detected unfilled interface skeletons and are counted as failures.

\noindent\textbf{Evaluated systems.}
We evaluate \NumSystems{} configurations from the Claude, GPT, Gemini, Qwen, GLM, and Kimi families \citep{anthropic2026opus5,openai2026gpt56sol,google2026gemini38flash,qwen2026qwen38,zai2026glm53,moonshot2026kimik3}.
Claude Code covers twelve configurations, while Codex and OpenClaw each cover GPT-5.6 Sol, GPT-5.5, and Qwen3.7-Max.
We retain each harness's native behavior because it is part of the deployed agent system.

\noindent\textbf{Reporting protocol.}
Scores use a $0$--$100$ scale and aggregate by task; the all-task average assigns zero to failures, while Avg.\ succ. excludes them.
Appendix~\ref{app:human_alignment} reports atomic agreement and holistic human validation on generated reports.
Appendices~\ref{app:evaluation}, \ref{app:experiments}, and~\ref{app:prompts} give evaluation protocols, configurations, budgets, detailed analyses, and prompts.

\subsection{Main Results}
\label{sec:main_results}

\begin{table*}[t]
\caption{Results on 50 tasks with task-bootstrap 95\% confidence intervals for Avg.
Avg. counts failures as zero, while other quality scores use successful reports.
Bold and underline mark the top two values except Success.}
\label{tab:main_results}
\centering
\scriptsize
\setlength{\tabcolsep}{3.0pt}
\resizebox{\textwidth}{!}{
\begin{tabular}{llrrrrrrrrr}
\toprule
\textbf{Harness} & \textbf{Model} & \textbf{Avg.} & \textbf{95\% CI} & \textbf{Success} & \textbf{Avg. succ.} & \textbf{Max} & \textbf{Min} & \textbf{Content} & \textbf{Presentation} & \textbf{Citation} \\
\midrule
\rowcolor{claudecodebg}& Claude Opus 5 & \textbf{63.3} & [57.8, 68.3] & 96.0\% & \textbf{66.0} & \underline{88.6} & 26.5 & \textbf{59.9} & \textbf{90.3} & \textbf{59.9} \\
\rowcolor{claudecodebg}& Claude Opus 4.8 & 60.7 & [56.8, 64.2] & 98.0\% & 62.0 & 81.8 & 37.2 & 57.4 & 86.0 & 53.5 \\
\rowcolor{claudecodebg}& Claude Opus 4.7 & 58.4 & [54.1, 62.1] & 96.0\% & 60.9 & 77.7 & \underline{44.5} & \underline{58.7} & 84.8 & 48.5 \\
\rowcolor{claudecodebg}& GPT-5.6 Sol & 54.9 & [52.9, 57.1] & 100.0\% & 54.9 & 74.3 & 39.6 & 42.0 & 85.7 & 56.0 \\
\rowcolor{claudecodebg}& GPT-5.5 & 50.4 & [48.0, 52.9] & 100.0\% & 50.4 & 74.0 & 28.7 & 38.7 & 80.9 & 49.6 \\
\rowcolor{claudecodebg}& Gemini-3.8-Flash & 52.6 & [50.2, 55.1] & 100.0\% & 52.6 & 73.2 & 28.3 & 43.6 & 79.6 & 49.7 \\
\rowcolor{claudecodebg}& Qwen3.8-Omni-Flash & 59.3 & [55.7, 62.5] & 98.0\% & 60.5 & 78.4 & \textbf{44.6} & 54.6 & 89.0 & 51.5 \\
\rowcolor{claudecodebg}& Qwen3.8-Max & 52.9 & [46.5, 58.8] & 88.0\% & 60.1 & \textbf{91.5} & 37.4 & 50.4 & \underline{89.5} & \underline{56.5} \\
\rowcolor{claudecodebg}& Qwen3.7-Max & 39.9 & [34.7, 44.9] & 88.0\% & 45.4 & 65.3 & 26.0 & 39.2 & 73.3 & 37.1 \\
\rowcolor{claudecodebg}& Qwen3.5-Omni-Plus & 28.1 & [21.8, 34.5] & 62.0\% & 45.4 & 64.8 & 32.9 & 39.9 & 71.1 & 37.4 \\
\rowcolor{claudecodebg}& GLM-5.3 & \underline{62.0} & [59.4, 64.6] & 100.0\% & \underline{62.0} & 79.1 & 38.0 & 58.2 & 83.2 & 54.4 \\
\rowcolor{claudecodebg}\multirow{-12}{*}{Claude Code} & Kimi-K3 & 50.7 & [45.4, 55.5] & 92.0\% & 55.1 & 75.1 & 31.0 & 48.1 & 82.0 & 48.8 \\
\rowcolor{codexbg}& GPT-5.6 Sol & 53.9 & [50.8, 57.0] & 100.0\% & 53.9 & 76.3 & 30.9 & 42.4 & 87.6 & 50.6 \\
\rowcolor{codexbg}& GPT-5.5 & 46.0 & [42.7, 49.2] & 98.0\% & 46.9 & 72.5 & 32.7 & 34.8 & 83.6 & 42.8 \\
\rowcolor{codexbg}\multirow{-3}{*}{Codex} & Qwen3.7-Max & 45.3 & [41.9, 48.6] & 100.0\% & 45.3 & 72.0 & 23.5 & 38.5 & 69.8 & 40.3 \\
\rowcolor{openclawbg}& GPT-5.6 Sol & 41.7 & [35.4, 47.6] & 82.0\% & 50.9 & 72.4 & 33.5 & 39.9 & 83.9 & 47.2 \\
\rowcolor{openclawbg}& GPT-5.5 & 41.2 & [36.2, 45.7] & 90.0\% & 45.8 & 71.1 & 24.3 & 33.9 & 81.9 & 41.4 \\
\rowcolor{openclawbg}\multirow{-3}{*}{OpenClaw} & Qwen3.7-Max & 35.3 & [29.8, 40.6] & 82.0\% & 43.0 & 73.4 & 18.3 & 37.2 & 70.2 & 34.5 \\
\bottomrule
\end{tabular}}
\end{table*}

\noindent\textbf{Overall performance.}
Table~\ref{tab:main_results} reports the main scores and their task-bootstrap intervals, while Appendix~\ref{app:score_robustness} tests scoring sensitivity.
Claude Opus 5 has the highest point estimate at $63.3$ overall and $66.0$ on successful reports, but misses $4.0\%$ of tasks.
GLM-5.3 completes every task and scores $62.0$.
Their $1.3$ point difference has a task-bootstrap $95\%$ confidence interval of $[-4.0,5.6]$, so their order is uncertain.
Qwen3.8-Max scores $60.1$ on successful reports, yet its $88.0\%$ completion lowers the overall score to $52.9$.

\noindent\textbf{Shared bottleneck.}
Table~\ref{tab:main_results} and Appendix Table~\ref{tab:research_requirements} show a shared gap between presentation and research substance.
Under the shared interface template, presentation averages $81.8$ points, compared with $45.4$ for content and $47.8$ for citation.
Across configurations, the unweighted mean of configuration-level direct-video-fact pass rates is $79.9\%$.
The mean falls to $8.7\%$ for external evidence, $8.8\%$ for cross-source verification, and $5.0\%$ for multi-fact composition.
Fact coverage, causal analysis, and comparison average only $7.6\%$, $9.5\%$, and $4.6\%$.
The dashboard additionally pools atomic decisions across configurations; under that weighting, the corresponding L0--L3 rates are $80.4\%$, $9.0\%$, $9.2\%$, and $5.0\%$, and the three coverage-direction rates are $7.9\%$, $9.7\%$, and $4.6\%$.
Video faithfulness averages $92.4\%$, but logical consistency and claim support reach only $21.2\%$ and $15.9\%$.
Inspection of failed coverage criteria shows that omissions of audited research targets dominate explicit contradictions.
\textbf{Reports satisfy presentation requirements more consistently than audited coverage, reasoning, and claim-support criteria.}

\noindent\textbf{Different systems fail differently.}
Under matched Claude Code runs, Table~\ref{tab:main_results} shows distinct strengths and failure modes.
Claude Opus 5 leads all three dimensions, while GLM-5.3 ranks second overall with complete delivery.
GLM-5.3 combines complete delivery with strong content and citation.
Qwen3.8-Omni-Flash is strong in content and presentation but weaker in citation, while Qwen3.8-Max has the highest single-report score but lower completion reliability.
GPT-5.6 Sol and Gemini-3.8-Flash complete every task but remain content-limited.
Kimi-K3 is mid-range, while Qwen3.5 and Qwen3.7 lag mainly in content, citation, and completion.

\noindent\textbf{Finding sources is easier than using them.}
Appendix Table~\ref{tab:citation_behaviors} exposes a recurring citation failure in the fine-grained directions.
For example, under Claude Code, GPT-5.6 Sol passes $98.0\%$ of source-count criteria and $93.5\%$ of multimodal-material criteria, but only $23.9\%$ of claim-support criteria.
Gemini-3.8-Flash passes $93.0\%$ of multimodal-material criteria, yet reaches $13.9\%$ on claim support.
Trace inspection shows that reports often collect links in a final Sources section without citing them beside the claims they should support.
Thus, retrieval volume does not imply evidence integration or citation closure between inline citations and source records.

\noindent\textbf{Video categories and query types.}
The four video families span only $2.0$ points, from $52.8$ to $54.8$ (Table~\ref{tab:family_results}), while subcategories range from $46.0$ to $59.1$.
General, video-view, and user-demand queries average $41.3$, $52.5$, and $50.4$ points.
These differences remain descriptive because subcategories contain one to five tasks and query forms are not randomly assigned; Appendix~\ref{app:category_results} provides the detailed breakdowns.

\begin{table}[t]
\caption{Successful-report means by major video family on a $0$ to $100$ scale.}
\label{tab:family_results}
\centering
\footnotesize
\setlength{\tabcolsep}{2.4pt}
\begin{tabular*}{\columnwidth}{@{\extracolsep{\fill}}lrrrr@{}}
\toprule
Family & Overall & Content & Citation & Presentation \\
\midrule
Knowledge and viewpoints & \textbf{54.8} & \textbf{48.9} & 46.9 & 81.2 \\
Consumer and business & 52.8 & 41.5 & \textbf{51.3} & \textbf{83.3} \\
Physical procedures & 53.5 & 43.7 & 50.2 & 83.1 \\
Creative and digital & 53.4 & 47.0 & 45.5 & 81.4 \\
\bottomrule
\end{tabular*}
\end{table}

\subsection{Further Analysis}
\label{sec:further_analysis}

\noindent\textbf{Skill ablations.}
Table~\ref{tab:ablation_results} and Appendix Table~\ref{tab:ablation_paired} separate atomic skill access from high-level orchestration.
The five-level ladder runs from an interface-only baseline to the full video-research skill, with exact capabilities and controls in Appendix~\ref{app:ablation}.
All variants use the same model, harness, iteration limit, timeout, retry policy, and report contract.
Actual compute use is determined by the agent under each capability configuration.

\begin{table*}[t]
\caption{Cumulative skill ablation on 15 matched tasks with GPT-5.6 Sol and Claude Code.
Columns abbreviate interface (UI), multimodal retrieval (Multi.), video-research orchestration (VDR), and presentation (Pres.).
Scores include failures except Avg. succ.}
\label{tab:ablation_results}
\centering
\scriptsize
\setlength{\tabcolsep}{2.4pt}
\begin{tabular*}{\textwidth}{@{\extracolsep{\fill}}lccccc rrrrrrrrr@{}}
\toprule
& \multicolumn{5}{c}{Provided capability} & \multicolumn{9}{c}{Evaluation result} \\
\cmidrule(lr){2-6}\cmidrule(lr){7-15}
Variant & UI & Video & Web & Multi. & VDR & Avg. & Success & Avg. succ. & Content & Pres. & Citation & Steps & Time (h) & Tokens (M) \\
\midrule
(i) & \cmark & & & & & 28.0 & 66.7\% & 42.0 & 18.9 & 50.5 & 28.2 & 95.5 & 0.6 & 1.2 \\
(ii) & \cmark & \cmark & & & & 31.8 & 73.3\% & 43.4 & 26.6 & 54.3 & 25.4 & 94.0 & 0.8 & 1.4 \\
(iii) & \cmark & \cmark & \cmark & & & 45.5 & 100.0\% & 45.5 & 34.9 & 80.3 & \underline{40.0} & 104.0 & 0.6 & 2.0 \\
(iv) & \cmark & \cmark & \cmark & \cmark & & \underline{46.6} & 100.0\% & \underline{46.6} & \underline{40.1} & \underline{80.7} & 34.7 & 103.0 & 0.7 & 2.3 \\
(v) & \cmark & \cmark & \cmark & \cmark & \cmark & \textbf{53.9} & 100.0\% & \textbf{53.9} & \textbf{40.9} & \textbf{83.9} & \textbf{55.4} & 167.0 & 1.6 & 4.7 \\
\bottomrule
\end{tabular*}
\end{table*}

Adding text web research to source-video processing raises the all-task mean from $31.8$ to $45.5$ points and success from $73.3\%$ to $100.0\%$.
Multimodal retrieval adds $1.1$ points overall: its $5.2$-point content gain is partly offset by a $5.3$-point citation loss, separating access to richer evidence from successful citation integration.
Full orchestration adds $7.3$ points over all atomic skills and improves 12 of 15 tasks, with a paired $95\%$ interval of $[3.4,11.1]$.
Its largest gain is $20.7$ citation points, while content rises by $0.8$ points: \textbf{the workflow's clearest benefit is improved citation quality}.
\looseness=-1

This benefit raises median steps from $103$ to $167$, time from $0.7$ to $1.6$ hours, and tokens from $2.3$M to $4.7$M, describing a quality--cost trade-off.

\noindent\textbf{Harness effects.}
Appendix~\ref{app:harness_results} and Table~\ref{tab:harness_summary} compare the three harnesses under matched models.
Claude Code and Codex both average $48.4$ points on the three matched models, but Codex completes $99.3\%$ of runs compared with $96.0\%$ for Claude Code.
Claude Code leads on both GPT models, while Codex leads on Qwen3.7-Max and has no median tool error.
OpenClaw uses fewer tools and tokens but has the lowest mean of $39.4$ and the lowest completion rate of $84.7\%$, showing that harness effects depend on the model and resource use.

\noindent\textbf{Research trajectories and efficiency.}
Appendix Tables~\ref{tab:trajectory_statistics} and~\ref{tab:trajectory_order} report phase order across all 18 configurations.
Under Claude Code, median post-write search ranges from $33.3\%$ for GPT-5.6 Sol to $61.5\%$ for Claude Opus 5, while GLM-5.3 reaches $54.2\%$ with $15.5$ median write blocks.
Eight configurations meet our interleaving rule, suggesting that drafting often exposes evidence gaps and triggers retrieval \citep{wang2026trajectoryerrors}.
Interleaving alone does not ensure quality because productive evidence transfer matters more than trajectory length.

\noindent\textbf{Failure patterns.}
Appendices~\ref{app:trajectory_cases} and~\ref{app:examples} show that short traces leave evidence gaps, while long traces may repeat errors or fail to transfer evidence into the report.
Fragmented traces can still recover, so steps, tokens, and tool errors alone do not predict quality.

\section{Related Work}
\label{sec:related_work}

\noindent\textbf{Deep research agents.}
Text agents produce cited reports through long-horizon retrieval and synthesis \citep{qiao2025webresearcher,li2025websailor,du2026openseeker}.
Multimodal report agents add visual retrieval and interleaved output \citep{geng2025webwatcher,shi2026vidr,ma2026tvir}, whereas video deep research connects source-video grounding with open-web search \citep{gao2026videosearcher,liu2026videorover,zhang2026adavdr}.
The former does not center research on a supplied video, and the latter mainly targets short-answer QA.
Our task uses the audiovisual timeline as the research backbone and returns a cited interactive report enriched with external multimodal evidence.

\noindent\textbf{Deep research benchmarks.}
Long-video and browsing benchmarks test temporally distributed understanding or open-web multi-step QA \citep{fu2024videomme,meng2026videozero,liang2025videobrowser,liu2026videodr}.
Text report benchmarks assess factuality, completeness, citation quality, and user intent \citep{du2025deepresearchbench,sharma2025researchrubrics,zhong2026draco}.
Multimodal, process, and rubric studies further examine visual retrieval, tool use, trajectory errors, report logic, and citation support \citep{huang2026mmdeepresearch,ye2026miroeval,wang2026trajectoryerrors,wan2026deepverifier,leung2026citationverifier}.
OmniDR-Bench instead jointly evaluates video grounding, multimodal evidence, claim support, and interaction in complete reports.

\section{Conclusion}
\label{sec:conclusion}

We introduce Video-Centered Deep Research Report Generation and OmniDR-Bench.
Omni-ART and ARC provide fine-grained evaluation grounded in audited multimodal evidence.
Across \NumSystems{} configurations, the best score is \BestAllTaskScore{} out of 100, exposing gaps in coverage, cross-modal reasoning, and claim support.
Future work targets better audiovisual use, efficiency, reasoning, and citations (Appendix~\ref{app:limitations}).

% Fill the last main-text page using the template's normal stretchable spacing.
\pagebreak
\section*{AI Use Statement}

Generative AI assisted with language polishing and writing revision.
AI models were also used as evaluated systems and as components of the benchmark evaluator.

\section*{Ethics Statement}

OmniDR-Bench uses public videos and web sources, which raise privacy, copyright, contextual integrity, and representation concerns.
We exclude private or unsafe requests and follow applicable source licenses and platform terms.
The set is not population representative, and its reports should not replace professional advice.
Human verification follows applicable review, consent, and data protection requirements.

\section*{Reproducibility Statement}

Sections~\ref{sec:task} through~\ref{sec:experiments} and the appendix specify the task, workflow, benchmark, evaluation, experiments, taxonomy, queries, rubrics, and prompts.
The release adds artifacts from S0 to S6, results, trajectories, evaluator code, harness configurations, scripts, and versioned source metadata.

\bibliography{iclr2027_conference}

@inproceedings{fu2024videomme,
  title={Video-mme: The first-ever comprehensive evaluation benchmark of multi-modal llms in video analysis},
  author={Fu, Chaoyou and Dai, Yuhan and Luo, Yongdong and Li, Lei and Ren, Shuhuai and Zhang, Renrui and Wang, Zihan and Zhou, Chenyu and Shen, Yunhang and Zhang, Mengdan and others},
  booktitle={2025 IEEE/CVF Conference on Computer Vision and Pattern Recognition (CVPR)},
  pages={24108--24118},
  year={2025},
  organization={IEEE}
}

@article{meng2026videozero,
  title={VideoZeroBench: Probing the Limits of Video MLLMs with Spatio-Temporal Evidence Verification},
  author={Meng, Jiahao and Tan, Yue and Xu, Qi and Wang, Haochen and Ren, Zhongwei and Liu, Weisong and Wang, Yuhao and Zhang, Renrui and Li, Xiangtai and Duan, Haodong and Tong, Yunhai and Yang, Ming-Hsuan},
  journal={arXiv preprint arXiv:2604.01569},
  year={2026}
}

@inproceedings{du2025deepresearchbench,
  title={DeepResearch Bench: A Comprehensive Benchmark for Deep Research Agents},
  author={Du, Mingxuan and Xu, Benfeng and Zhu, Chiwei and Zhang, Licheng and Wang, Xiaorui and Mao, Zhendong},
  booktitle={International Conference on Learning Representations},
  year={2026}
}

@article{sharma2025researchrubrics,
  title={ResearchRubrics: A Benchmark of Prompts and Rubrics for Evaluating Deep Research Agents},
  author={Sharma, Manasi and Zhang, Chen Bo Calvin and Bandi, Chaithanya and Wang, Clinton and Aich, Ankit and others},
  journal={arXiv preprint arXiv:2511.07685},
  year={2025}
}

@article{liang2025videobrowser,
  title={Video-browser: Towards agentic open-web video browsing},
  author={Liang, Zhengyang and Shu, Yan and Liu, Xiangrui and Qin, Minghao and Liang, Kaixin and Sebe, Nicu and Liu, Zheng and Liao, Lizi},
  journal={arXiv preprint arXiv:2512.23044},
  year={2025}
}

@article{liu2026videodr,
  title={Watching, Reasoning, and Searching: A Video Deep Research Benchmark on Open Web for Agentic Video Reasoning},
  author={Liu, Chengwen and Yu, Xiaomin and Chang, Zhuoyue and Huang, Zhe and Zhang, Shuo and others},
  journal={arXiv preprint arXiv:2601.06943},
  year={2026}
}

@article{huang2026mmdeepresearch,
  title={MMDeepResearch-Bench: A Benchmark for Multimodal Deep Research Agents},
  author={Huang, Peizhou and Zhong, Zixuan and Wan, Zhongwei and Zhou, Donghao and Alam, Samiul and others},
  journal={arXiv preprint arXiv:2601.12346},
  year={2026}
}

@inproceedings{an2026aeseval,
  title={Can Vision Language Models Assess Graphic Design Aesthetics? A Benchmark, Evaluation, and Dataset Perspective},
  author={An, Arctanx and Sun, Shizhao and Huang, Danqing and Cheng, Mingxi and Gao, Yan and Li, Ji and Qiao, Yu and Bian, Jiang},
  booktitle={International Conference on Learning Representations},
  year={2026},
  eprint={2603.01083},
  archivePrefix={arXiv},
  primaryClass={cs.CV}
}

@article{ye2026miroeval,
  title={MiroEval: Benchmarking Multimodal Deep Research Agents in Process and Outcome},
  author={Ye, Fangda and Hu, Yuxin and Zhu, Pengxiang and Li, Yibo and Jin, Ziqi and others},
  journal={arXiv preprint arXiv:2603.28407},
  year={2026}
}

@article{du2026openseeker,
  title={OpenSeeker-v2: Pushing the Limits of Search Agents with Informative and High-Difficulty Trajectories},
  author={Du, Yuwen and Ye, Rui and Tang, Shuo and Huang, Keduan and Zhu, Xinyu and Cai, Yuzhu and Chen, Siheng},
  journal={arXiv preprint arXiv:2605.04036},
  year={2026}
}

@article{zhang2026ptah,
  title={Towards Verifiable Multimodal Deep Research: A Multi-Agent Harness for Interleaved Report Generation},
  author={Zhang, Chenghao and Dong, Guanting and Liu, Yufan and Zhao, Tong and Li, Xiaoxi and Dou, Zhicheng},
  journal={arXiv preprint arXiv:2605.29861},
  year={2026}
}

@article{ma2026tvir,
  title={{TVIR}: Building Deep Research Agents Towards Text-Visual Interleaved Report Generation},
  author={Ma, Xinkai and Bai, Zhiqi and Zhang, Dingling and Liu, Pei and Yuan, Yishuo and others},
  journal={arXiv preprint arXiv:2606.02320},
  year={2026}
}

@article{gao2026videosearcher,
  title={VideoSearcher: Empowering Video Deep Research with Multi-Tool Agentic Reasoning via Reinforcement Learning},
  author={Gao, Zhenkun and Bao, Yicheng and Peng, Jinlong and Li, Xueheng and Huang, Theo and others},
  journal={arXiv preprint arXiv:2607.02927},
  year={2026}
}

@article{fang2026videodeepresearch,
  title={Video-DeepResearch: Towards the Next-Generation Multimodal Deepresearch Agent},
  author={Fang, Zhen and Zeng, Yu and Huang, Wenxuan and Zhao, Yiming and Huang, Shiting and others},
  journal={arXiv preprint arXiv:2608.03979},
  year={2026}
}

@article{liu2026videorover,
  title={Thinking Beyond Videos: Unifying Video Reasoning and Deep Research for Open-World Video Agents},
  author={Liu, Wenqi and Ma, Shijie and Wang, Yunxiao and Liu, Meng and Su, Qile and others},
  journal={arXiv preprint arXiv:2608.23329},
  year={2026}
}

@article{zhang2026adavdr,
  title={{AdaVDR}: Adaptive Tool Use and Reflection for Video Deep Research},
  author={Zhang, Xintong and Fan, Xiaomeng and Yan, Shilin and He, Ekko and Liu, Zicheng and others},
  journal={arXiv preprint arXiv:2608.25559},
  year={2026}
}

@article{java2025characterizing,
  title={Characterizing Deep Research: A Benchmark and Formal Definition},
  author={Java, Abhinav and Khandelwal, Ashmit and Midigeshi, Sukruta and Halfaker, Aaron and Deshpande, Amit and Goyal, Navin and Gupta, Ankur and Natarajan, Nagarajan and Sharma, Amit},
  journal={arXiv preprint arXiv:2508.04183},
  year={2025}
}

@article{qiao2025webresearcher,
  title={WebResearcher: Unleashing Unbounded Reasoning Capability in Long-Horizon Agents},
  author={Qiao, Zile and Chen, Guoxin and Chen, Xuanzhong and Yu, Donglei and Yin, Wenbiao and Wang, Xinyu and Zhang, Zhen and Li, Baixuan and Yin, Huifeng and Li, Kuan and others},
  journal={arXiv preprint arXiv:2509.13309},
  year={2025}
}

@article{li2025websailor,
  title={WebSailor: Navigating Super-human Reasoning for Web Agent},
  author={Li, Kuan and Zhang, Zhongwang and Yin, Huifeng and Zhang, Liwen and Ou, Litu and Wu, Jialong and Yin, Wenbiao and Li, Baixuan and Tao, Zhengwei and Wang, Xinyu and others},
  journal={arXiv preprint arXiv:2507.02592},
  year={2025}
}

@article{geng2025webwatcher,
  title={WebWatcher: Breaking New Frontier of Vision-Language Deep Research Agent},
  author={Geng, Xinyu and Xia, Peng and Zhang, Zhen and Wang, Xinyu and Wang, Qiuchen and Ding, Ruixue and Wang, Chenxi and Wu, Jialong and Zhao, Yida and Li, Kuan and others},
  journal={arXiv preprint arXiv:2508.05748},
  year={2025}
}

@article{wan2026deepverifier,
  title={Inference-Time Scaling of Verification: Self-Evolving Deep Research Agents via Test-Time Rubric-Guided Verification},
  author={Wan, Yuxuan and Fang, Tianqing and Li, Zaitang and Huo, Yintong and Wang, Wenxuan and Mi, Haitao and Yu, Dong and Lyu, Michael R.},
  journal={arXiv preprint arXiv:2601.15808},
  year={2026}
}

@article{zhong2026draco,
  title={{DRACO}: A Cross-Domain Benchmark for Deep Research Accuracy, Completeness, and Objectivity},
  author={Zhong, Joey and Zhang, Hao and Southern, Clare and Yang, Jeremy and Wang, Thomas and Jung, Kate and Zhang, Shu and Yarats, Denis and Ho, Johnny and Ma, Jerry},
  journal={arXiv preprint arXiv:2602.11685},
  year={2026}
}

@article{zhao2026reportlogic,
  title={ReportLogic: Evaluating Logical Quality in Deep Research Reports},
  author={Zhao, Jujia and Huan, Zhaoxin and Wang, Zihan and Zhang, Xiaolu and Zhou, Jun and Verberne, Suzan and Ren, Zhaochun},
  journal={arXiv preprint arXiv:2602.18446},
  year={2026}
}

@article{shi2026vidr,
  title={{ViDR}: Grounding Multimodal Deep Research Reports in Source Visual Evidence},
  author={Shi, Zhuofan and Jia, Peilun and Sun, Baoqin and Shen, Haiyang and Xie, Sixiong and Ma, Yun and Jing, Xiang},
  journal={arXiv preprint arXiv:2605.13034},
  year={2026}
}

@article{wang2026trajectoryerrors,
  title={Where Do Deep-Research Agents Go Wrong? Span-Level Error Localization in Agent Trajectories},
  author={Wang, Jiaming and Feng, Ziteng and Wu, Jiangtao and Li, Ruihao and Xie, Qianqian and Ren, Yuxiang and Zhu, He and Han, Xueming and Meng, Fanyu and Feng, Junlan and Liu, Jiaheng},
  journal={arXiv preprint arXiv:2606.02060},
  year={2026}
}

@article{leung2026citationverifier,
  title={Do You Need a Frontier Model as a Citation Verifier? Benchmarking Rubric LLMs for Deep-Research Source Attribution},
  author={Leung, Ethan and Lumer, Elias and Feld, Corey and Huber, Austin and Subbiah, Vamse Kumar and Paul, Kevin},
  journal={arXiv preprint arXiv:2607.08700},
  year={2026}
}

@misc{anthropic2026opus5,
  title={Introducing {Claude Opus 5}},
  author={{Anthropic}},
  year={2026},
  url={https://www.anthropic.com/news/claude-opus-5}
}

@misc{anthropic2026opus48,
  title={Introducing {Claude Opus 4.8}},
  author={{Anthropic}},
  year={2026},
  url={https://www.anthropic.com/news/claude-opus-4-8}
}

@misc{anthropic2026opus47,
  title={Introducing {Claude Opus 4.7}},
  author={{Anthropic}},
  year={2026},
  url={https://www.anthropic.com/news/claude-opus-4-7}
}

@misc{openai2026gpt56sol,
  title={Previewing {GPT-5.6 Sol}},
  author={{OpenAI}},
  year={2026},
  url={https://openai.com/index/previewing-gpt-5-6-sol/}
}

@misc{openai2026gpt55,
  title={Introducing {GPT-5.5}},
  author={{OpenAI}},
  year={2026},
  url={https://openai.com/index/introducing-gpt-5-5/}
}

@misc{qwen2026qwen37,
  title={{Qwen3.7}},
  author={{Qwen Team}},
  year={2026},
  url={https://qwen.ai/blog?id=qwen3.7}
}

@misc{qwen2026qwen38,
  title={{Qwen3.8}},
  author={{Qwen Team}},
  year={2026},
  url={https://qwen.ai/blog?id=qwen3.8}
}

@misc{qwen2026qwen38omniflash,
  title={{Qwen3.8-Omni-Flash}},
  author={{Qwen Team}},
  year={2026},
  url={https://qwen.ai/blog?id=qwen3.8-omni-flash}
}

@misc{qwen2026qwen35omni,
  title={{Qwen3.5-Omni}},
  author={{Qwen Team}},
  year={2026},
  url={https://qwen.ai/blog?id=qwen3.5-omni}
}

@misc{google2026gemini38flash,
  title={{Gemini 3.8 Flash} Model Card},
  author={{Google DeepMind}},
  year={2026},
  url={https://deepmind.google/models/model-cards/gemini-3-8-flash/}
}

@misc{zai2026glm53,
  title={{GLM-5.3}},
  author={{Z.AI}},
  year={2026},
  url={https://z.ai/blog/glm-5.3}
}

@misc{moonshot2026kimik3,
  title={{Kimi K3}},
  author={{Moonshot AI}},
  year={2026},
  url={https://www.kimi.com/news/kimi-k3}
}
\bibliographystyle{iclr2027_conference}

\clearpage
\appendix
\onecolumn
\section*{Appendix}

\noindent\textbf{Contents.}
The appendix provides the complete evaluation specification, task taxonomy, construction protocol, experimental details, prompts, examples, and limitations.

\begin{itemize}
\setlength{\itemsep}{2pt}
\setlength{\parskip}{0pt}
\item Appendix~\ref{app:evaluation}: Complete evaluation specification.
\item Appendix~\ref{app:taxonomy}: Task taxonomy and collection details.
\item Appendix~\ref{app:queries}: Query construction details.
\item Appendix~\ref{app:workflow}: Detailed agent workflow.
\item Appendix~\ref{app:construction}: Benchmark construction and annotation.
\item Appendix~\ref{app:experiments}: Additional experimental details.
\item Appendix~\ref{app:prompts}: Prompt and agent protocols.
\item Appendix~\ref{app:examples}: Qualitative task examples.
\item Appendix~\ref{app:limitations}: Limitations and future work.
\end{itemize}

\suppressfloats[t]

\section{Complete Evaluation Specification}
\label{app:evaluation}

\subsection{Released Hyperparameters}
\label{app:hyperparameters}

The method uses symbols in its definitions so that the evaluation rule remains separate from one released configuration.
Table~\ref{tab:released_hyperparameters} records the current values used to produce the reported scores.
All values are fixed before evaluating the compared systems.

\begin{table*}[t]
\caption{Symbols and current settings for construction, execution, and evaluation.}
\label{tab:released_hyperparameters}
\centering
\scriptsize
\setlength{\tabcolsep}{3.4pt}
\begin{tabular}{p{0.18\textwidth}p{0.53\textwidth}p{0.21\textwidth}}
\toprule
Symbol & Role & Released setting \\
\midrule
$\boldsymbol{\alpha}^{(v)}$ & Video-view query ranking weights & $\VideoQueryWeights$ \\
$\boldsymbol{\alpha}^{(u)}$ & User-demand query ranking weights & $\UserQueryWeights$ \\
$T_{\mathrm{win}}$ & Source-video analysis window & $\VideoWindowSeconds$ seconds \\
$B_{\mathrm{web}}, B_{\mathrm{video}}$ & Minimum external-source and related-video budgets & $\MinExternalSources, \MinRelatedVideos$ \\
$B_{\mathrm{doc}}, B_{\mathrm{anchor}}$ & Minimum document and per-anchor evidence budgets & $\MinDocuments, \MinSourcesPerAnchor$ \\
$\tau_d$ & Saturation for each high-volume coverage direction & $\CoverageSaturation$ \\
$\kappa_d$ & Atom-count cap for each high-volume direction & $\DirectionCountCap$ \\
$(\lambda_{C},\lambda_{R},\lambda_{P})$ & Content, citation, and presentation weights & $(\ContentWeight,\CitationWeight,\PresentationWeight)$ \\
$(N_{\mathrm{vote}},T_J)$ & Number of model-judge votes for gate criteria and judge temperature & $(\JudgeVotes,\JudgeTemperature)$ \\
$R_{\max}$ & Maximum issue-driven repair rounds & $\MaxRepairRounds$ \\
$\delta_{\mathrm{cal}}$ & Minimum score gap from $R_1$ to $R_3$ & $\CalibrationGap$ \\
\bottomrule
\end{tabular}
\end{table*}

\subsection{Omni-ART Direction Inventory}

\begin{figure*}[t]
  \centering
  \includegraphics[width=0.99\textwidth]{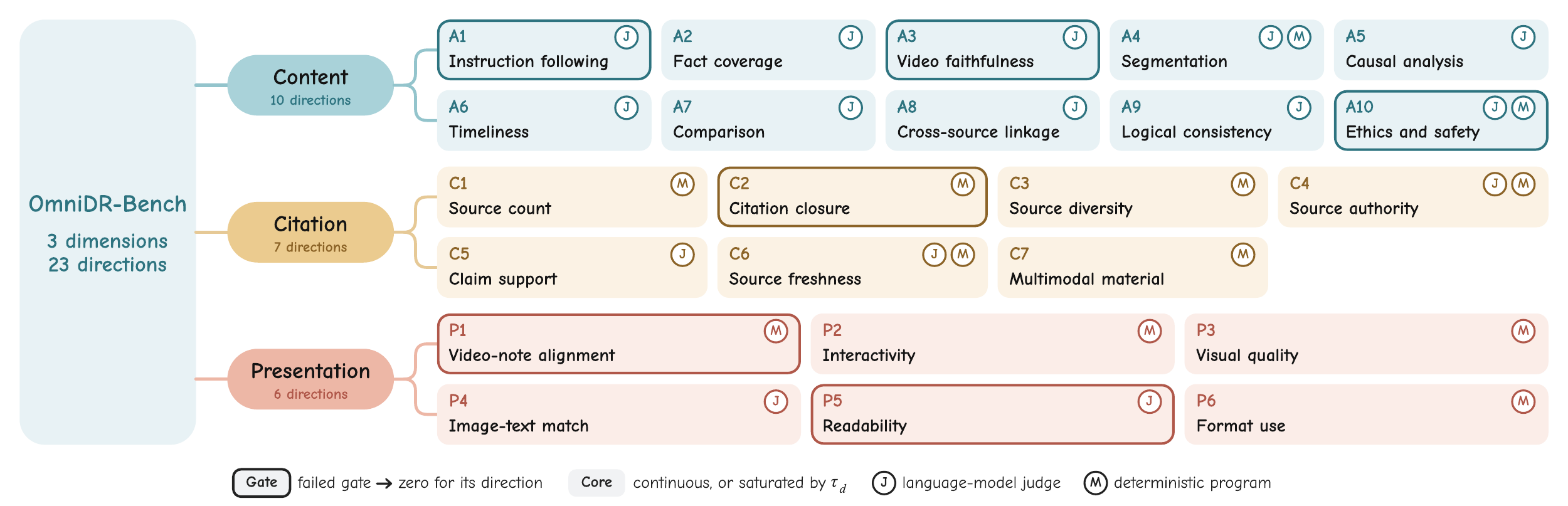}
  \caption{Omni-ART hierarchy.
  Three dimensions and 23 directions test audiovisual grounding and cross-modal reasoning, source use and claim support, and interactive presentation.
  Task-specific binary leaves use a language-model judge (J), a deterministic program (M), or both, while outlined directions are gates.}
  \label{fig:rubric_tree}
\end{figure*}

The evaluator uses fixed direction definitions and instance-specific Omni-ART leaves.
Figure~\ref{fig:rubric_tree} gives the complete direction inventory.
The evaluator column indicates the check type.
J is a multimodal model judgment.
M is a deterministic verification, including static program rules and browser interaction probes.
Figure~\ref{fig:rubric_tree} also distinguishes gate directions from core diagnostics.
A failed gate zeros its direction.
The three high-volume directions use saturation $\tau_d$, while the remaining core directions aggregate continuously.

All content directions are primarily judged with J, except observable structural conditions that can be checked with M.
The mixed design reserves model judgments for semantic properties and uses deterministic checks whenever the artifact exposes a direct signal.

\subsection{Model-Judge Protocol}

\noindent\textbf{Judge choice.}
GPT-5.6 Sol can inspect both text and images, which is required for semantic content, citation-support, and visual-presentation criteria.
It also supports bilingual instructions and structured outputs over long report contexts.
We use the same model and prompt settings for every evaluated system.

\noindent\textbf{Decision protocol.}
The judge receives one atomic binary criterion, its evidence note, and the relevant report material.
It returns a binary decision, a short rationale, and supporting report evidence when available.
It runs at temperature zero.
Each non-gate model criterion is judged once.
Gate criteria are evaluated three times and decided by majority vote because a failed gate sets its direction score to zero.
M criteria evaluate directly observable properties without a model judge.
Their deterministic checks include browser interaction probes.
\noindent\textbf{Interpretation.}
Fixing the judge improves comparability across systems but does not remove model-specific bias.
The human study in Appendix~\ref{app:human_protocol} verifies benchmark facts and criteria rather than every automatic report-level decision.
The separate study in Appendix~\ref{app:human_alignment} measures agreement between automatic and human rubric decisions on generated reports.
Scores should therefore be interpreted under the released judge, prompts, and evaluator version.

\subsection{Scoring and Failure Handling}
\label{app:na_guard}

For task $i$, direction $d$, and applicable atom $j$, the binary outcome is $a_{i,d,j}$.
Let $n_{i,d}$ be the number of applicable atoms for task $i$ and direction $d$.
The observed direction rate is
\begin{equation}
r_{i,d}=\frac{\sum_j a_{i,d,j}}{n_{i,d}}.
\end{equation}
Fact coverage, causal analysis, and comparison are high-volume directions.
Their credited rates use $c_{i,d}=\min(1,r_{i,d}/\tau_d)$, while other directions use $c_{i,d}=r_{i,d}$.
The saturation constant reflects the amount of required evidence that can fit in the standard report budget and is fixed before system comparison.
The released value follows the report evidence budget divided by the median retained fact volume.
Thus, the aggregate is a budget-normalized rubric score; raw atom pass rates separately report coverage of the audited evidence targets.

Each direction also evaluates its declared logic.
For a core direction, this logic produces only a diagnostic status and does not change its credited rate.
A failed gate receives an effective rate of zero for that direction.
The default rule does not erase its complete dimension.
The within-dimension weight is the number of scored atoms, capped by $\kappa_d$ for C-FC, C-CA, and C-CMP.
The cap prevents these high-volume directions from dominating the content dimension.
The final score uses weights $\lambda_C$, $\lambda_R$, and $\lambda_P$ for content, citation, and presentation.
The released weights prioritize research substance, then source support, while retaining a smaller presentation component.
Appendix~\ref{app:score_robustness} tests alternative values without rerunning the judge.

\noindent\textbf{Guarded citation N/A.}
The intended preference is a correctly stated and cited fact, followed by a correctly stated fact without adequate citation, followed by an omitted fact.
A failed citation atom becomes an N/A candidate only when its linked, positive-polarity coverage checks have at least one scored decision and all scored decisions are zero.
Failure to explain a mechanism or solve a reasoning problem alone is not evidence of omission.
Passing citation atoms retain their credit.

Removing a failed citation atom also shrinks the citation denominator, so eligibility alone does not authorize N/A propagation.
Before final aggregation, the implementation applies a direction-level guard.
Let $\widehat w_d^C$ and $\widehat w_j^R$ be normalized within-dimension weights using the same count caps as Equation~\ref{eq:score_aggregation}.
For an affected content direction $d$, let $n_d$ be its scored atom count and $\gamma_d$ its marginal-gain factor: $1/\tau_d$ below coverage saturation, zero at saturation, and one for an ordinary direction without saturation.
For citation direction $j$, let $n_j$ and $r_j$ be the scored count and pass rate before the candidate removals, and $k_j$ the number of candidates still proposed for removal.
With $\mathcal{B}$ denoting content directions linked to these candidates, the guard compares
\begin{equation}
\label{eq:na_guard}
\widehat\Delta_C
=\min_{d\in\mathcal{B}}\lambda_C\widehat w_d^C\frac{\gamma_d}{n_d},
\qquad
\widehat\Delta_R
=\lambda_R\sum_{j:n_j>k_j}\widehat w_j^R r_j\frac{k_j}{n_j-k_j}.
\end{equation}
These quantities are the direction-level content marginal estimate and citation-denominator gain estimate used by the implementation.
N/A assignments are retained only while $\widehat\Delta_C>\widehat\Delta_R$.
Otherwise, the evaluator restores one candidate to its original zero score, choosing from the fact with the most proposed removals, and recomputes the guard until it holds or no candidates remain.
The final applicable sets and their weights are then used in Equations~\ref{eq:direction_rates}--\ref{eq:score_aggregation}.

At coverage saturation, $\gamma_d=0$ triggers rollback: omitting a fact cannot receive the proposed denominator bonus merely because both reports already obtain full coverage credit.
The stated and omitted cases can tie at saturation, so the intended preference does not imply a strict score increase beyond the cap.
The evaluator records candidate, applied, and retracted counts and both guard readings in \texttt{na\_guard}; the offline \texttt{na-order} check verifies the guard on saved evaluations.
Evaluator failures are separately marked as skipped and excluded from their direction denominators.
A missing or unreadable report contributes zero to the all-task average.

\subsection{Human Verification and Agreement}
\label{app:human_protocol}

Human verification checks the two benchmark artifacts that define each task.
Fact-library review asks whether every retained fact is supported by the cited video moment or external source.
Criterion review asks whether every criterion is correct, judgeable, and bound to sufficient evidence.
Reviewers inspect the complete fact library and complete criterion set for each assigned task.

The process contains two iterative validation rounds over a shrinking candidate pool.
In each round, multiple reviewers independently inspect every assigned package in the current pool.
Each fact and criterion is marked correct or incorrect under the supplied evidence and annotation rules.
The item correctness rate is the percentage of reviewed fact and criterion judgments marked correct.
Unanimity is the fraction of reviewed items for which all three reviewers assign the same label.
S6 aggregates these item judgments into package-level error profiles after each round.
Repairable packages are revised, packages with high or persistent artifact error rates are removed, and only accepted packages proceed to the next round.
This human quality-control loop reduces the candidate pool from 82 to 60 and then to the final 50 tasks.
Table~\ref{tab:human_consistency} reports item-level rates before the repair and filtering decisions prompted by each round; the second-round value is not a post-correction error estimate for the released snapshot.
After final correction, reviewers sample 20 fact or criterion items from each of the 50 task packages.
More than $94.0\%$ of these 1,000 sampled items are marked correct.

\begin{table}[h]
\caption{Two-round human verification during task-package curation.
Percentages are rounded to one decimal.}
\label{tab:human_consistency}
\centering
\footnotesize
\setlength{\tabcolsep}{3.2pt}
\begin{tabular*}{\columnwidth}{@{\extracolsep{\fill}}lrr@{}}
\toprule
Round & Correctness & Unanimity \\
\midrule
1 & \HumanRoundOneAccuracy{} & \HumanRoundOneUnanimity{} \\
2 & \HumanRoundTwoAccuracy{} & \HumanRoundTwoUnanimity{} \\
\bottomrule
\end{tabular*}
\end{table}

\subsection{Human Alignment on Generated Reports}
\label{app:human_alignment}

We separately check the reliability of automatic rubric evaluation on generated reports.
We randomly sample 15 reports generated by Claude Opus 4.8 and 15 generated by GPT-5.6 Sol.
Three human raters independently evaluate each report under the released Omni-ART criteria.
They also assign a holistic score from 1 to 5 based on the report's usefulness and trustworthiness.
Inter-rater agreement divides the decisions with three matching human labels by all comparable decisions.
Machine-to-human agreement divides automatic decisions that match the human majority label by all comparable decisions.
The two agreement rates are $93.0\%$ and $91.0\%$, respectively.
Across all 50 tasks, Claude Opus 4.8 and GPT-5.6 Sol score $60.7$ and $54.9$ with Omni-ART.
On the 15 sampled reports per model, their mean holistic ratings are $3.73$ and $3.51$, respectively.
Across the 30 reports, Omni-ART scores correlate with mean holistic ratings at Spearman $\rho=0.64$.
Table~\ref{tab:human_alignment} summarizes both atomic agreement and holistic validation.

\begin{table}[h]
\caption{Human validation on 30 randomly sampled reports.
Holistic ratings use a 1--5 scale, and the reported correlation uses report-level scores.}
\label{tab:human_alignment}
\centering
\footnotesize
\setlength{\tabcolsep}{3.2pt}
\begin{tabular*}{\columnwidth}{@{\extracolsep{\fill}}lrr@{}}
\toprule
Measure & Reports & Result \\
\midrule
Human unanimity on atoms & 30 & 93.0\% \\
Machine-human atom agreement & 30 & 91.0\% \\
Opus 4.8 Omni-ART score & 50 & 60.7/100 \\
GPT-5.6 Sol Omni-ART score & 50 & 54.9/100 \\
Opus 4.8 holistic rating & 15 & 3.73/5 \\
GPT-5.6 Sol holistic rating & 15 & 3.51/5 \\
Omni-ART vs. holistic rating & 30 & $\rho=0.64$ \\
\bottomrule
\end{tabular*}
\end{table}

\section{Task Taxonomy and Collection Details}
\label{app:taxonomy}

\subsection{Scenario Taxonomy}

The taxonomy starts from user needs rather than video genres.
It asks why a viewer would need research after watching and which audiovisual evidence makes the source video necessary.
Table~\ref{tab:full_taxonomy} lists the four families, all 19 subcategories, and their typical research needs.

\begin{table*}[t]
\caption{Full scenario taxonomy and typical research needs.
One video may support multiple queries.}
\label{tab:full_taxonomy}
\centering
\scriptsize
\setlength{\tabcolsep}{3.2pt}
\begin{tabular}{p{0.20\textwidth}p{0.27\textwidth}p{0.47\textwidth}}
\toprule
Family & Subcategory & Typical research need \\
\midrule
\multirow{5}{0.20\textwidth}{Knowledge and viewpoints}
& Academic frontier & Recover prerequisites, trace papers, check newer methods, and assess reproducibility \\
& Macro economy & Verify statistics, update policy context, and trace effects to industries or firms \\
& Law and traffic & Connect visible events to rules, precedent, evidence requirements, and uncertainty \\
& Controversy and history & Check claims against records and compare competing accounts \\
& Hardcore science & Explain mechanisms, test boundary conditions, and identify safe reproduction settings \\
\midrule
\multirow{4}{0.20\textwidth}{Consumer and business}
& Product review & Verify product claims, compare alternatives, and assess long-term failure evidence \\
& Company financials & Reconcile commentary with filings, current results, and industry comparisons \\
& Livestream commerce & Check advertising claims, qualifications, prices, and comparable products \\
& B2B software & Compare integration, migration, customization, cost, and operational risk \\
\midrule
\multirow{5}{0.20\textwidth}{Physical procedures}
& Life and DIY & Recover steps, parameters, substitutions, and failure diagnosis \\
& Industrial equipment & Identify components and construct safe operation or maintenance procedures \\
& Agricultural machinery & Adapt equipment and parameters to terrain, weather, and load conditions \\
& Lab precision & Recover exact settings, calibration, safety constraints, and troubleshooting paths \\
& Fitness and rehabilitation & Check movement, loading, contraindications, and person-specific adaptations \\
\midrule
\multirow{5}{0.20\textwidth}{Creative and digital}
& Design and 3D editing & Reproduce effects across versions, plugins, parameters, and reusable assets \\
& Coding and automation & Recover code, dependencies, commands, and error handling \\
& Gaming strategy & Reconstruct mechanics and update tactics for patches, equipment, or failure points \\
& UI and dashboards & Recover hidden operations, simplify effects, and adapt templates \\
& AI tools and workflow & Reconstruct prompts and APIs, then check stability, cost, and current documentation \\
\bottomrule
\end{tabular}
\end{table*}

The taxonomy covers three recurring user goals.
The first is efficient understanding, where the report locates important moments and explains difficult material.
The second is verification and extension, where external evidence checks claims and supplies missing context or alternatives.
The third is practical application, where the report adapts a procedure, recommendation, or decision to a concrete need.
These goals may coexist within one task.

\subsection{Video Inclusion and Exclusion}

\noindent\textbf{Inclusion criteria.}
A candidate video must contain several researchable claims, entities, procedures, parameters, or unresolved questions.
At least one important element must depend on visual, audiovisual, or temporal evidence rather than a title or transcript alone.
The topic must also have enough public external evidence for verification, comparison, explanation, updating, or application.

\noindent\textbf{Boundary tests.}
Each candidate is checked with the three acceptance tests in Table~\ref{tab:boundary_tests_full}.
The tests are applied to the complete video and query pair rather than to the video title alone.
For example, the pendant-light task depends on the shown socket, coupling, assembly, and product markings, while its safety assessment requires external standards and product evidence (Table~\ref{tab:qualitative_tasks}).
These checks establish video dependence during curation; Table~\ref{tab:ablation_results} studies capability groups and does not separately quantify the contribution of frames versus transcripts.

\begin{table}[t]
\caption{Acceptance tests used during task selection.}
\label{tab:boundary_tests_full}
\centering
\scriptsize
\setlength{\tabcolsep}{3.2pt}
\begin{tabular}{p{0.27\columnwidth}p{0.65\columnwidth}}
\toprule
Test & Acceptance condition \\
\midrule
Video-only & The request cannot be completed from the source video alone and requires external information \\
Web-only & Generic web research cannot replace the given video because replacing the source changes the investigation and at least one important anchor requires visual or temporal evidence \\
Answerability & Public evidence is sufficient within a practical research budget \\
\bottomrule
\end{tabular}
\end{table}

\noindent\textbf{Exclusion criteria.}
We reject direct lookups, transcript restatements, queries whose answer is independent of the source video, and cases with insufficient public evidence.
We also reject tasks that require private information, create avoidable safety risks, or depend on sources that cannot be stably identified.

\subsection{Fixed Data Snapshot}

The current benchmark contains 50 tasks and \VideoHours{} hours of source-video content.
It includes 23 Chinese tasks and 27 English tasks drawn from YouTube and Bilibili.
The query set contains 7 general queries, 17 video-view queries, and 26 user-demand queries.
The four families are near-balanced, while subcategory findings remain descriptive because some scenarios have few tasks.

\section{Query Construction Details}
\label{app:queries}

\subsection{Three Query Forms}

We use three query forms to separate broad video understanding from focused research needs.
Their roles and final counts are shown in Table~\ref{tab:query_forms}.

\begin{table}[t]
\caption{Query forms in the fixed 50-task set.}
\label{tab:query_forms}
\centering
\scriptsize
\setlength{\tabcolsep}{3.0pt}
\begin{tabular}{p{0.19\columnwidth}rp{0.62\columnwidth}}
\toprule
Form & Count & Intended scope \\
\midrule
General & 7 & A complete video-centered research report \\
Video-view & 17 & A broad research direction grounded in the video \\
User-demand & 26 & A concrete decision, diagnosis, reproduction, or application need \\
\bottomrule
\end{tabular}
\end{table}

A general query requests broad research over the complete source video.
A video-view query selects one theory, case, claim, object, style, or procedure that is present in the video and asks for wider investigation.
A user-demand query starts from a plausible viewer's concrete objective and uses the video as necessary evidence for that objective.
The two focused views may share a video anchor, but they differ in who defines the research boundary.

\subsection{Candidate Mining Pipeline}

\noindent\textbf{Step 1: audiovisual understanding.}
The video is probed for duration and language, divided into windows of length $T_{\mathrm{win}}$ when needed, and processed by parallel video readers.
Timestamped speech, visible text, events, objects, and uncertain regions are merged into one dense understanding record.
This record is the only shared input to the two query miners.

\noindent\textbf{Step 2: independent query views.}
An anchor miner proposes video-view questions from the theories, claims, cases, visual objects, and procedures present in the record.
A demand miner uses the scenario playbook to propose realistic user-demand questions.
The demand miner considers six user roles, namely novice, practitioner, reproducer, decision maker, verifier, and troubleshooter.

\noindent\textbf{Step 3: scoring and deduplication.}
Each miner returns five to ten candidates with a rationale and a confidence score.
Near duplicates within and across views are removed before ranking.
The video-view score is $(\boldsymbol{\alpha}^{(v)})^\top\mathbf{s}^{(v)}$.
Here, $\boldsymbol{\alpha}^{(v)}$ is the fixed weight vector and $\mathbf{s}^{(v)}$ contains video centrality, research value, searchability, and distinctness.
The user-demand score is $(\boldsymbol{\alpha}^{(u)})^\top\mathbf{s}^{(u)}$.
Here, $\boldsymbol{\alpha}^{(u)}$ is the fixed weight vector and $\mathbf{s}^{(u)}$ contains video necessity, external research need, concreteness, and demand fit.
The released vectors are listed in Appendix~\ref{app:hyperparameters}.

\noindent\textbf{Step 4: schema and language checks.}
The candidates are sorted within each view and written to a fixed JSON schema.
The output language follows the primary spoken language of the source video.
Automated checks verify the required keys, candidate count, score range, ranking order, and language consistency.

\subsection{Manual Acceptance Protocol}

Annotators review the highest-ranked candidates against the full audiovisual record and the category-specific demand playbook.
A video-view query is retained only when it is broad enough to require synthesis, specific enough to remain centered on the source, and supported by searchable external evidence.
A user-demand query is retained only when three conditions hold together.
It must express a user's own task, depend on details from the given video, and require substantial external research.

The strip-the-persona test detects demand queries that only attach a user label to a generic research topic.
Annotators remove the persona phrase and inspect the remainder.
If the remainder only asks whether the video's claim is correct, the candidate is reclassified or rejected.
If it still describes a concrete action, decision, diagnosis, or reproduction problem, it remains a user-demand query.

We reject candidates that merely ask for the next step in the video or request one isolated fact.
We also reject search recipes without a user need and requests that remain unchanged when the source video is swapped.
The final choice balances query quality with language, platform, family, and subcategory coverage.

\subsection{Representative Query Transformations}

Table~\ref{tab:query_examples} illustrates the difference among the three forms.
The focused examples are translated when the original task is Chinese.

\begin{table*}[t]
\caption{Representative query forms and the evidence that keeps each query video-centered.}
\label{tab:query_examples}
\centering
\scriptsize
\setlength{\tabcolsep}{3.2pt}
\begin{tabular}{p{0.13\textwidth}p{0.48\textwidth}p{0.31\textwidth}}
\toprule
Form & Query & Why the source video is necessary \\
\midrule
General & Conduct a general, in-depth, video-centered research analysis of this video. & The complete timeline supplies the candidate claims, demonstrations, and visual anchors \\
Video-view & Study the trade-off between under-display-camera image quality and display quality, centered on the device shown in the video. & The shown device, tests, and visual artifacts define the object and comparison boundary \\
Video-view & Study the distinction between the equivalence point and the reaction endpoint in the titration curve explained by the video. & The demonstrated curve, terminology, and procedure determine what must be explained and checked \\
User-demand & Assess whether the video's claim about low debt-to-asset ratios among high-income households supports taking a leveraged home loan under current policy and market conditions. & The claim and its original measurement frame come from the video, while the decision needs current policy and external data \\
\bottomrule
\end{tabular}
\end{table*}

\section{Detailed Agent Workflow}
\label{app:workflow}

\subsection{Roles and Shared State}

The main agent owns the user goal, execution order, shared evidence state, report assembly, and repair decisions.
Specialized \texttt{video-reader} and \texttt{doc-reader} agents isolate long audiovisual and document contexts and return structured records.
Each record stores its modality, source, date, excerpt, associated research anchor, and verification state.

\subsection{Seven Execution Stages}

\noindent\textbf{Stage 1: acquire and inspect the video.}
The workflow resolves the video source, probes its duration, frame rate, resolution, and audio, analyzes the request, and prepares an execution plan.

\noindent\textbf{Stage 2: build the audiovisual timeline.}
The video is divided into windows of $T_{\mathrm{win}}$ seconds and assigned to parallel \texttt{video-reader} instances.
Each reader analyzes visual content and speech, and the main agent merges the results under absolute timestamps.

\noindent\textbf{Stage 3: select research anchors.}
The main agent selects request-relevant claims, entities, procedures, and parameters from the timeline.
Each anchor retains its timestamp, transcript evidence, or frame crop and states whether the goal is to explain, compare, verify, update, or apply the video content.

\noindent\textbf{Stage 4: research external multimodal evidence.}
Anchor-specific branches search webpages, images, videos, and documents in parallel.
The workflow opens sources beyond snippets, assigns related videos to a \texttt{video-reader}, and assigns long or visual documents to a \texttt{doc-reader}.
Each branch returns evidence records with provenance, excerpts, media, dates, and confidence.

\noindent\textbf{Stage 5: connect evidence and synthesize.}
The workflow relates video and external evidence through support, contradiction, update, explanation, comparison, and application.
Corrective and comparative findings require independent support before the main agent converts them into report claims.

\noindent\textbf{Stage 6: write data and render the report.}
The agent incrementally writes the overview, timeline segments, research findings, media, and sources into a structured \texttt{window.DATA} object.
After schema and asset checks, a fixed renderer creates a dual-pane report with the source video, time-aligned notes, evidence, citations, and clickable timestamps.

\noindent\textbf{Stage 7: validate, repair, and deliver.}
Programmatic checks validate report data, local assets, citations, rendering, and browser interactions.
Fresh-context evidence checks reopen cited sources and test whether they support the associated claims.
Required-process checks confirm inspection of related videos, use of visual evidence, and verification of key data.
The main agent repairs unsupported claims, replaces unreachable evidence, reruns affected checks, and delivers the validated report.

\subsection{Released Budgets}

The reference configuration uses $B_{\mathrm{web}}$ and $B_{\mathrm{video}}$ for minimum external-source and related-video counts.
It uses $B_{\mathrm{doc}}$ and $B_{\mathrm{anchor}}$ for minimum document and per-anchor source counts.
Table~\ref{tab:released_hyperparameters} reports their released values.

\section{Benchmark Construction and Annotation}
\label{app:construction}

\subsection{Seven-Stage Construction and Human Curation}

Audited Rubric Construction (ARC) turns each accepted video--query pair into a validated task package through seven stages, S0--S6.
The stages form an iterative quality-control loop: S0--S4 construct the task artifacts, S5 detects failures through automatic and human validation, and S6 either repairs the package or removes it before the next round.
Table~\ref{tab:construction_artifacts} summarizes this flow.

\begin{table*}[t]
\caption{The seven ARC stages, their artifacts, and their quality controls.}
\label{tab:construction_artifacts}
\centering
\scriptsize
\setlength{\tabcolsep}{3.0pt}
\begin{tabular}{lp{0.22\textwidth}p{0.28\textwidth}p{0.37\textwidth}}
\toprule
Stage & Operation & Main output & Quality control \\
\midrule
S0 & Specify the task & Task scope and search plan & Source-video dependence, research focus, language, metadata, and route coverage \\
S1 & Build the fact library & Timestamped video evidence, external evidence, and reconciled facts & Isolated collection, provenance, conflict preservation, relevance filtering, and semantic deduplication \\
S2 & Generate candidate criteria & Evidence-bound atomic questions & Binding to fact records or executable report properties \\
S3 & Screen the criteria & Validated criteria with difficulty labels & Grounding, atomicity, objectivity, judgeability, uniqueness, and coverage \\
S4 & Assemble Omni-ART & A task-specific rubric tree & Three dimensions, 23 directions, and per-direction caps \\
S5 & Validate the task package & Static, behavioral, stability, and human audit results & Checks of facts and criteria plus package-level issue rates \\
S6 & Repair, filter, and freeze & A revised package or an exclusion decision & Route repairable issues to their source stage; remove packages with pervasive errors; advance accepted packages to the next round \\
\bottomrule
\end{tabular}
\end{table*}

S0--S4 first produce \NumCandidates{} candidate task packages.
During S5, reviewers inspect the facts and criteria in each package and record incorrect, unsupported, ambiguous, or unjudgeable items.
S6 aggregates these findings at the package level: repairable issues are routed back to the stage that produced them, while packages with high or persistent artifact error rates are excluded rather than carried into evaluation.
Only accepted or successfully repaired packages proceed to the next validation round.
The first human-validation and S6 pass reduces the pool from 82 to 60 packages; the next pass yields the final 50-task benchmark.
Thus, 82-to-60-to-50 is the attrition trace of the S5--S6 human quality-control loop, not a separate selection based on scores obtained by compared systems.

\subsection{Independent Evidence Acquisition}

ARC separates video reading from external retrieval so that neither branch can copy the other's framing.
The two evidence maps are reconciled only after each branch records its own observations and provenance.

\noindent\textbf{Video evidence map.}
S1a runs without web access and records timestamped speech, visible text, entities, actions, procedures, parameters, and unresolved visual details.
Long videos are split into windows of length $T_{\mathrm{win}}$ for parallel audiovisual reading.
Every retained video fact has a time reference and its origin is marked as video.

\noindent\textbf{Web evidence map.}
S1b has no path to the source video and searches six routes in isolated sessions.
The routes cover named entities, primary provenance, concepts and standards, relevant people, analogous cases, and event timelines.
Retrieval includes webpages, documents, images, and related videos because visual demonstrations and comparisons may not be recoverable from text alone.
Each route records atomic claims, canonical URLs, source type, publication date, and a supporting quotation.
Discovered entities are fed into later rounds.
Collection stops after two consecutive saturated rounds or at the configured round limit.
Saturation requires both new relevant claims and new relevant source domains to fall below their thresholds.

\noindent\textbf{Reconciliation.}
S1c merges semantically equivalent statements while retaining alternate wording and provenance.
Numerical or factual disagreements are preserved as explicit conflict edges rather than silently resolved.
Confidence is assigned by evidence structure.
A record is verified when it has at least two independent domains or a suitable primary source.
It is single-source when it has one source, disputed when supported records conflict, and unavailable when no usable source is found.

The evidence record is the auditable interface between retrieval and rubric construction.
Table~\ref{tab:evidence_record} shows the information preserved for each atomic claim.
These fields ground later Omni-ART criteria in sources, video locations, verification states, relevance labels, and unresolved conflicts.

\begin{table}[t]
\caption{Evidence-record schema for generating and auditing source-bound Omni-ART criteria.}
\label{tab:evidence_record}
\centering
\scriptsize
\setlength{\tabcolsep}{3.0pt}
\begin{tabular}{lp{0.63\columnwidth}}
\toprule
Field & Meaning \\
\midrule
\texttt{fid} & Stable fact identifier \\
\texttt{claim} & One checkable statement \\
\texttt{type} & Numeric, entity, date, place, quote, mechanism, opinion, or structural \\
\texttt{origin} & Video, external, or both \\
\texttt{video\_ref} & Source-video time interval when applicable \\
\texttt{sources} & URL, quotation, date, and provenance tier such as primary, official, academic, media, or secondary \\
\texttt{confidence} & Verified, single-source, disputed, or unavailable \\
\texttt{relevance} & Core, background, or unrelated \\
\texttt{conflicts\_with} & Identifiers of contradictory records \\
\bottomrule
\end{tabular}
\end{table}

\subsection{Rubric Tree Leaf Generation and Screening}

S2 maps evidence types to candidate templates for fact coverage, faithfulness, causal explanation, comparison, cross-source linkage, citation support, and report behavior.
This mapping ensures that the rubric tests both source-video understanding and the use of external multimodal evidence.
Each model-judged atom contains a fact identifier, and each machine atom without a fact identifier states the reader-facing impact of failure.
Criteria are phrased as binary questions with a fixed expected polarity.
Compound statements and important numbers are split into separate atoms.

S3 removes candidates that lack evidence, contain several claims, use subjective language, duplicate another item, admit several interpretations, or fall outside their assigned direction.
Difficulty is computed from evidence structure rather than predicted by a model.
L0 requires one source-video fact, L1 requires external evidence, L2 requires independent verification or conflict handling, and L3 requires composition across distinct facts.

S4 places the retained criterion leaves into stable direction slots.
An absent direction must have an instance-specific N/A reason.
Caps prevent high-volume directions from overwhelming the remaining evaluation criteria.
The fixed snapshot contains 4,060 content atoms, 1,234 citation atoms, and 1,120 presentation atoms.
It contains 2,363 L0, 1,475 L1, 2,167 L2, and 409 L3 atoms.
The retained rubrics reference \NumFacts{} task-local fact identifiers, averaging approximately \MeanFactsPerTask{} per task.

\subsection{Validation at Three Levels and Human Review}

\noindent\textbf{Static validation.}
Sixteen checks cover schema validity, identifiers, direction membership, logic compatibility, evidence binding, gates, and presentation impact.
They also test machine-check eligibility, subjective wording, duplicates, direction coverage, and difficulty balance.
The remaining checks examine source concentration, empty-report behavior, direction boundaries, and language consistency.
Errors must be resolved before release, while warnings require an explicit review decision.
The empty-report check runs every deterministic atom against a valid but content-free report and rejects criteria that award unsupported credit.

\noindent\textbf{Behavioral validation.}
$R_1$ receives the video but cannot search the web.
$R_2$ uses Claude Opus 4.8 under Claude Code with the full tool set and video-research skill.
$R_3$ receives compact evidence-map briefs.
The same model and harness generate all three reports so that the calibration varies information access rather than the report generator.
This is a construction-time check of sensitivity to evidence access; report-level agreement with human rubric decisions is evaluated separately in Appendix~\ref{app:human_alignment}.
Each task must satisfy $R_1<R_2<R_3$ and $R_3-R_1>\delta_{\mathrm{cal}}$, where $\delta_{\mathrm{cal}}$ is the minimum calibration gain.
All \NumTasks{} retained tasks pass both conditions.
Each $R_2$ report is also scored three times to test judge stability during calibration.
Across tasks, only $0.89\%$ of applicable criteria receive non-unanimous decisions on average.
This repeated scoring is an S5 stability test and is separate from the released evaluation protocol.

\noindent\textbf{Human validation.}
Reviewers inspect the facts and criteria in each assigned candidate package.
Each package is independently checked by multiple reviewers in the current validation pool.
For each fact and criterion, reviewers mark whether it is correct under the provided source evidence and annotation rules.
Majority vote determines each item label, and three-reviewer unanimity measures consistency.
Round one records \HumanRoundOneAccuracy{} item correctness and \HumanRoundOneUnanimity{} unanimity.
S6 uses the resulting package-level error profiles to repair viable packages and exclude high-error packages before the next round.
Round two records \HumanRoundTwoAccuracy{} item correctness and \HumanRoundTwoUnanimity{} unanimity on the surviving pool.
Its feedback drives the final repair and filtering pass before the snapshot is frozen.
Appendix~\ref{app:human_protocol} reports the protocol and results.

\noindent\textbf{Issue-driven repair.}
Fact errors return to evidence acquisition, while criterion errors return to criterion generation or rubric assembly.
After at most $R_{\max}$ revision rounds, packages that pass validation are frozen with their evidence and evaluator versions; packages with persistent or pervasive errors are excluded.

\section{Additional Experimental Details}
\label{app:experiments}

\subsection{System Configurations and Budgets}

The compared models and harnesses are listed in Table~\ref{tab:main_results}.
Version-specific sources document Claude Opus 5, 4.8, and 4.7 \citep{anthropic2026opus5,anthropic2026opus48,anthropic2026opus47}, GPT-5.6 Sol and GPT-5.5 \citep{openai2026gpt56sol,openai2026gpt55}, Gemini-3.8-Flash \citep{google2026gemini38flash}, Qwen3.8-Omni-Flash, Qwen3.8-Max, Qwen3.7-Max, and Qwen3.5-Omni-Plus \citep{qwen2026qwen38omniflash,qwen2026qwen38,qwen2026qwen37,qwen2026qwen35omni}, GLM-5.3 \citep{zai2026glm53}, and Kimi-K3 \citep{moonshot2026kimik3}.
All systems receive the same report contract and tool environment.
Agents may inspect the source video, search webpages and multimodal sources, and test the rendered report in a browser.
Each report attempt has a 64K-token budget and a limit of 500 agent iterations.
The process tables use different accounting units: steps count logged trajectory entries, while token totals accumulate usage across model requests, including repeated prompt processing.
A failed model request may be retried up to \MaxRequestRetries{} times.
Each task receives at most \MaxReportAttempts{} complete report attempts, with a \ReportTimeoutHours{}-hour limit for each attempt.
If all attempts fail or time out, the task is counted as unsuccessful.
A run otherwise succeeds only when its final HTML report is recoverable and can be parsed by the evaluator.

\subsection{Statistical Uncertainty and Score Sensitivity}
\label{app:score_robustness}

We estimate uncertainty by resampling the 50 task identities with replacement for 20,000 bootstrap replicates.
All systems are kept paired within each replicate, and failed tasks retain a score of zero.
The resulting percentile intervals appear in Table~\ref{tab:main_results}.
For Claude Opus 5 minus GLM-5.3, the paired mean difference is $1.3$ points and its 95\% confidence interval is $[-4.0,5.6]$.
The interval includes zero, so their order should be interpreted as a point-estimate ranking rather than a reliable separation.

Table~\ref{tab:score_sensitivity} reaggregates the saved binary criterion decisions under alternative scoring parameters.
It changes one parameter group at a time and requires no new model judgments.
The alternatives retain the same top three systems in the same order.
No system moves by more than one rank among all 18 configurations.
Absolute scores vary most when the dimension weights or coverage saturation changes, so comparisons should retain the released parameter setting.

\begin{table*}[t]
\caption{Sensitivity of the all-task ranking to score aggregation parameters.
The final column is the largest absolute rank change from the released setting.}
\label{tab:score_sensitivity}
\centering
\scriptsize
\setlength{\tabcolsep}{3.2pt}
\begin{tabular*}{\columnwidth}{@{\extracolsep{\fill}}lccclc@{}}
\toprule
Setting & $(\lambda_C,\lambda_R,\lambda_P)$ & $\tau_d$ & $\kappa_d$ & Top three systems with scores & Max. rank change \\
\midrule
Released & $(0.50,0.30,0.20)$ & 0.18 & 15 & Opus 5 63.3, GLM-5.3 62.0, Opus 4.8 60.7 & 0 \\
Equal weights & $(0.33,0.33,0.33)$ & 0.18 & 15 & Opus 5 67.2, GLM-5.3 65.3, Opus 4.8 64.3 & 1 \\
Content heavy & $(0.60,0.25,0.15)$ & 0.18 & 15 & Opus 5 61.9, GLM-5.3 61.0, Opus 4.8 59.5 & 1 \\
Citation heavy & $(0.40,0.40,0.20)$ & 0.18 & 15 & Opus 5 63.3, GLM-5.3 61.7, Opus 4.8 60.3 & 1 \\
Lower saturation & $(0.50,0.30,0.20)$ & 0.12 & 15 & Opus 5 65.9, GLM-5.3 65.1, Opus 4.8 63.6 & 0 \\
Higher saturation & $(0.50,0.30,0.20)$ & 0.24 & 15 & Opus 5 60.5, GLM-5.3 59.4, Opus 4.8 57.9 & 1 \\
Lower count cap & $(0.50,0.30,0.20)$ & 0.18 & 10 & Opus 5 63.4, GLM-5.3 62.3, Opus 4.8 60.9 & 1 \\
Higher count cap & $(0.50,0.30,0.20)$ & 0.18 & 20 & Opus 5 63.2, GLM-5.3 62.0, Opus 4.8 60.6 & 0 \\
\bottomrule
\end{tabular*}
\end{table*}

\subsection{Direction-Level Error Analysis}

The direction analysis in Section~\ref{sec:main_results} covers all 835 evaluable reports from the 18 configurations.
For each report and direction, the pass rate is the number of passed atoms divided by the number of applicable atoms.
Table~\ref{tab:research_requirements} reports every configuration and their cross-system mean.
Its columns follow the L0 to L3 evidence levels and selected official Omni-ART directions.
The final row is the unweighted arithmetic mean of the 18 configuration-level rates; it is therefore distinct from the dashboard's pooled-atom aggregate, whose denominator weights configurations according to their numbers of scored atoms.

\begin{table*}[t]
\caption{Atomic pass rates for all 18 configurations and their arithmetic mean.
FC, CA, CMP, VF, LC, and CS denote fact coverage, causal analysis, comparison, video faithfulness, logical consistency, and claim support.}
\label{tab:research_requirements}
\centering
\scriptsize
\setlength{\tabcolsep}{2.7pt}
\resizebox{\textwidth}{!}{
\begin{tabular}{llrrrrrrrrrr}
\toprule
& & \multicolumn{4}{c}{Evidence level} & \multicolumn{5}{c}{Content directions} & \multicolumn{1}{c}{Citation} \\
\cmidrule(lr){3-6}\cmidrule(lr){7-11}\cmidrule(lr){12-12}
Harness & Model & L0 & L1 & L2 & L3 & FC & CA & CMP & VF & LC & CS \\
\midrule
Claude Code & Claude Opus 5 & 88.7 & 11.8 & 15.1 & 12.0 & 12.3 & 10.7 & 13.1 & 97.6 & 34.4 & 20.9 \\
Claude Code & Claude Opus 4.8 & 85.7 & 11.4 & 12.2 & 7.5 & 10.5 & 12.8 & 6.9 & 89.6 & 23.0 & 22.1 \\
Claude Code & Claude Opus 4.7 & 83.2 & 12.6 & 14.0 & 8.7 & 12.3 & 14.2 & 6.1 & 94.3 & 24.8 & 23.1 \\
Claude Code & GPT-5.6 Sol & 84.8 & 7.1 & 6.9 & 2.7 & 5.7 & 6.9 & 1.5 & 90.4 & 20.2 & 23.9 \\
Claude Code & GPT-5.5 & 80.0 & 6.6 & 4.9 & 1.5 & 4.5 & 5.2 & 1.5 & 89.0 & 16.9 & 22.9 \\
Claude Code & Gemini-3.8-Flash & 78.2 & 8.4 & 7.0 & 4.2 & 6.6 & 8.2 & 3.7 & 83.8 & 17.1 & 13.9 \\
Claude Code & Qwen3.8-Omni-Flash & 86.9 & 10.0 & 10.8 & 6.2 & 9.4 & 12.6 & 5.4 & 96.3 & 20.5 & 18.3 \\
Claude Code & Qwen3.8-Max & 87.9 & 10.3 & 10.5 & 7.6 & 8.8 & 10.5 & 8.7 & 93.9 & 23.9 & 18.4 \\
Claude Code & Qwen3.7-Max & 69.7 & 8.0 & 9.4 & 3.1 & 7.2 & 10.4 & 4.6 & 87.3 & 25.7 & 14.5 \\
Claude Code & Qwen3.5-Omni-Plus & 70.4 & 6.0 & 5.7 & 3.5 & 5.5 & 6.4 & 2.3 & 93.8 & 10.4 & 16.5 \\
Claude Code & GLM-5.3 & 84.3 & 11.8 & 13.5 & 5.2 & 11.3 & 12.9 & 7.6 & 94.9 & 18.7 & 22.7 \\
Claude Code & Kimi-K3 & 80.1 & 9.9 & 10.7 & 5.4 & 9.2 & 10.7 & 4.3 & 93.5 & 22.8 & 17.8 \\
\midrule
Codex & GPT-5.6 Sol & 84.7 & 7.2 & 8.0 & 2.4 & 7.0 & 8.7 & 3.2 & 94.4 & 17.2 & 9.5 \\
Codex & GPT-5.5 & 79.5 & 7.8 & 4.6 & 1.8 & 5.0 & 5.7 & 3.5 & 93.3 & 14.8 & 6.4 \\
Codex & Qwen3.7-Max & 69.4 & 7.2 & 7.1 & 5.3 & 7.2 & 6.0 & 3.3 & 88.9 & 19.0 & 7.3 \\
\midrule
OpenClaw & GPT-5.6 Sol & 80.7 & 7.9 & 6.5 & 4.7 & 5.1 & 13.5 & 1.9 & 93.9 & 23.6 & 10.5 \\
OpenClaw & GPT-5.5 & 77.4 & 5.6 & 5.3 & 3.5 & 4.3 & 8.2 & 2.3 & 96.8 & 20.3 & 6.1 \\
OpenClaw & Qwen3.7-Max & 66.7 & 7.5 & 6.7 & 4.7 & 5.7 & 7.7 & 3.8 & 91.2 & 29.2 & 11.7 \\
\midrule
\multicolumn{2}{l}{Mean across systems} & 79.9 & 8.7 & 8.8 & 5.0 & 7.6 & 9.5 & 4.6 & 92.4 & 21.2 & 15.9 \\
\bottomrule
\end{tabular}}
\end{table*}
We also group failed-criterion rationales into four outcomes.
The report may omit the fact, mention only its broad topic, omit its specific detail, or state an explicit factual error.
Omissions at different levels of specificity dominate explicit factual contradictions.
This analysis reuses the released atomic judgments and does not introduce an additional evaluator.
Table~\ref{tab:citation_behaviors} illustrates why source acquisition and source use must be evaluated separately.
Both selected systems frequently include multimodal materials, yet support far fewer individual claims.

\begin{table}[t]
\caption{Selected citation-related atomic pass rates on the final task set.}
\label{tab:citation_behaviors}
\centering
\scriptsize
\setlength{\tabcolsep}{2.8pt}
\begin{tabular*}{\columnwidth}{@{\extracolsep{\fill}}lrrr@{}}
\toprule
Model & Source count & Multimodal material & Claim support \\
\midrule
GPT-5.6 Sol & 98.0\% & 93.5\% & 23.9\% \\
Gemini-3.8-Flash & 24.0\% & 93.0\% & 13.9\% \\
\bottomrule
\end{tabular*}
\end{table}

\subsection{Results by Video Category}
\label{app:category_results}

The category analysis uses the successful-report mean from the fixed 50-task comparison page.
Missing or unreadable reports are excluded from this diagnostic view because completion is reported separately in Table~\ref{tab:main_results}.
Table~\ref{tab:query_results} reports all-task means by query form, with failures assigned zero.
Table~\ref{tab:model_family_results} reports successful-report means for each system and major video family.
Table~\ref{tab:subcategory_results} reports all 19 subcategories.

\begin{table}[t]
\caption{Results by query form over all 18 systems.}
\label{tab:query_results}
\centering
\footnotesize
\setlength{\tabcolsep}{3.2pt}
\begin{tabular*}{\columnwidth}{@{\extracolsep{\fill}}lrr@{}}
\toprule
Query form & Tasks & Avg. \\
\midrule
General & 7 & 41.3 \\
Video-view & 17 & 52.5 \\
User-demand & 26 & 50.4 \\
\bottomrule
\end{tabular*}
\end{table}

\begin{table*}[t]
\caption{Successful-report means by model and major video family.
K, C, P, and D denote knowledge, consumer, physical, and digital videos.}
\label{tab:model_family_results}
\centering
\scriptsize
\setlength{\tabcolsep}{4.0pt}
\begin{tabular*}{\columnwidth}{@{\extracolsep{\fill}}llrrrr@{}}
\toprule
Harness & Model & K & C & P & D \\
\midrule
Claude Code & Claude Opus 5 & 68.1 & 69.9 & 64.3 & 62.6 \\
Claude Code & Claude Opus 4.8 & 62.0 & 60.8 & 64.0 & 60.7 \\
Claude Code & Claude Opus 4.7 & 63.6 & 60.0 & 63.0 & 56.8 \\
Claude Code & GPT-5.6 Sol & 57.1 & 54.1 & 53.1 & 55.0 \\
Claude Code & GPT-5.5 & 51.1 & 49.2 & 50.5 & 50.3 \\
Claude Code & Gemini-3.8-Flash & 53.9 & 51.7 & 50.9 & 53.5 \\
Claude Code & Qwen3.8-Omni-Flash & 62.6 & 55.3 & 62.4 & 60.2 \\
Claude Code & Qwen3.8-Max & 54.4 & 61.8 & 61.2 & 62.1 \\
Claude Code & Qwen3.7-Max & 46.0 & 43.9 & 41.5 & 49.2 \\
Claude Code & Qwen3.5-Omni-Plus & 43.7 & 49.0 & 43.7 & 45.4 \\
Claude Code & GLM-5.3 & 61.9 & 61.6 & 62.7 & 61.8 \\
Claude Code & Kimi-K3 & 57.2 & 58.1 & 51.6 & 53.8 \\
Codex & GPT-5.6 Sol & 53.9 & 52.4 & 54.6 & 54.3 \\
Codex & GPT-5.5 & 46.7 & 42.3 & 49.9 & 47.6 \\
Codex & Qwen3.7-Max & 44.8 & 45.8 & 43.9 & 46.7 \\
OpenClaw & GPT-5.6 Sol & 56.1 & 44.9 & 49.0 & 52.0 \\
OpenClaw & GPT-5.5 & 50.1 & 42.9 & 42.4 & 46.3 \\
OpenClaw & Qwen3.7-Max & 44.2 & 43.4 & 45.1 & 39.8 \\
\bottomrule
\end{tabular*}
\end{table*}

\begin{table*}[t]
\caption{Successful-report means for all video subcategories on a $0$ to $100$ scale.}
\label{tab:subcategory_results}
\centering
\scriptsize
\setlength{\tabcolsep}{3.4pt}
\begin{tabular*}{\columnwidth}{@{\extracolsep{\fill}}clrrrrr@{}}
\toprule
Family & Subcategory & Tasks & Overall & Content & Citation & Presentation \\
\midrule
Knowledge & Hardcore science & 2 & 46.0 & 28.3 & 51.8 & 81.4 \\
Physical & Fitness and rehabilitation & 3 & 46.0 & 27.5 & 51.5 & 84.1 \\
Physical & Agricultural machinery & 1 & 48.7 & 33.2 & 51.0 & 83.9 \\
Consumer & Livestream commerce & 3 & 50.2 & 37.4 & 49.8 & 82.8 \\
Consumer & B2B software & 2 & 50.6 & 38.3 & 51.0 & 80.7 \\
Digital & UI and dashboards & 4 & 50.6 & 40.1 & 48.3 & 80.3 \\
Knowledge & Academic frontier & 1 & 52.0 & 47.2 & 41.4 & 80.2 \\
Digital & AI tools and workflow & 3 & 52.7 & 41.0 & 50.9 & 84.7 \\
Digital & Coding and automation & 1 & 53.9 & 51.2 & 39.1 & 83.2 \\
Digital & Design and 3D editing & 4 & 53.9 & 47.8 & 46.7 & 80.0 \\
Consumer & Product review & 2 & 54.1 & 43.9 & 50.8 & 84.5 \\
Physical & Life and DIY & 4 & 55.3 & 47.7 & 49.8 & 82.5 \\
Knowledge & Controversy and history & 4 & 55.5 & 50.6 & 46.0 & 81.8 \\
Physical & Lab precision & 2 & 56.1 & 50.3 & 48.1 & 82.4 \\
Knowledge & Macro economy & 2 & 56.2 & 50.9 & 48.6 & 80.8 \\
Knowledge & Law and traffic & 5 & 57.7 & 55.4 & 46.0 & 80.9 \\
Consumer & Company financials & 2 & 57.9 & 48.8 & 54.7 & 85.6 \\
Physical & Industrial equipment & 3 & 58.3 & 53.0 & 50.8 & 83.0 \\
Digital & Gaming strategy & 2 & 59.1 & 66.6 & 32.2 & 80.7 \\
\bottomrule
\end{tabular*}
\end{table*}

The broad families differ by only 2.0 score points, while the subcategory means span 13.1 points.
The lowest subcategories combine dense factual claims, fast audiovisual presentation, and consequential parameters that require external checking.
The highest subcategories tend to expose clearer state variables, comparison targets, or procedural boundaries.
These interpretations are descriptive because each subcategory contains only one to five tasks.
They motivate a larger balanced study rather than a claim that topic alone determines difficulty.

\subsection{Controlled Skill Ablations}
\label{app:ablation}

The ablation suite fixes GPT-5.6 Sol, Claude Code, the report contract, and 15 source tasks.
It also fixes the iteration limit, timeout, and retry policy across all variants.
Actual steps, elapsed time, and token use remain outcomes of the agent's behavior under each capability configuration.
The frozen subset contains four tasks from each of the first three major families and three from the fourth.
It also contains ten YouTube videos and five Bilibili videos.
Evaluating every configuration on every source produces 75 paired runs.
Configuration order is rotated within each task to reduce time and changing-web confounds.
Table~\ref{tab:ablation_results} defines the five cumulative capability settings and reports their results.
The fixed interface template contains no procedural instructions.
Video processing includes transcription, frame extraction, and source-video multimodal reading.
Text web research adds search and webpage reading, while multimodal retrieval further adds image and video search, download, and inspection.
The video-research skill adds the high-level workflow for anchor discovery, multi-source investigation, synthesis, rendering, and validation.
The final two rows measure the effect of adding high-level orchestration when all atomic skills are already available, including the extra research and verification it induces.
They do not hold realized compute constant because the model decides how much of the fixed allowance to use.
The comparison therefore estimates the workflow's deployed quality and cost trade-off.

Quality is compared with paired task differences and bootstrap confidence intervals.
Bootstrap samples resample source identities and keep the five configurations for each source together.
The full workflow exceeds the all-atomic-skills variant by $7.3$ points on average and wins 12 of 15 paired tasks.
The corresponding paired bootstrap 95\% confidence interval is $[3.4, 11.1]$.
Completion is tested separately because a configuration may improve quality among successful reports while producing fewer valid reports.
Process metrics include wall time, runner-reported tokens, trajectory steps, tool errors, source consumption, and verifier activity.

\begin{table}[t]
\caption{Paired effect of adding the video-research workflow to all atomic skills.}
\label{tab:ablation_paired}
\centering
\footnotesize
\setlength{\tabcolsep}{4.0pt}
\begin{tabular*}{\columnwidth}{@{\extracolsep{\fill}}lrrr@{}}
\toprule
Comparison & Mean gain & Wins & 95\% CI \\
\midrule
Full vs. all atomic skills & 7.3 & 12/15 & [3.4, 11.1] \\
\bottomrule
\end{tabular*}
\end{table}

\subsection{Controlled Harness Comparison}
\label{app:harness_results}

Table~\ref{tab:harness_summary} compares the three harnesses only on the shared GPT-5.6 Sol, GPT-5.5, and Qwen3.7-Max models.
Quality and success average over the three matched model runs.
Process columns take the median of the three model-level medians over available task traces.
They use runner-native measurements and are not hardware-normalized costs.

\begin{table*}[t]
\caption{Controlled harness comparison over the three shared base models.
Steps, tool calls, and errors are counts, time is in hours, and tokens are in millions.}
\label{tab:harness_summary}
\centering
\scriptsize
\setlength{\tabcolsep}{4.0pt}
\begin{tabular*}{\columnwidth}{@{\extracolsep{\fill}}lrrrrrrr@{}}
\toprule
Harness & Avg. & Success & Steps & Tool calls & Tool errors & Time & Tokens \\
\midrule
Claude Code & 48.4 & 96.0\% & 156.0 & 105.5 & 7.0 & 1.3 & 4.3 \\
Codex & 48.4 & 99.3\% & 134.0 & 114.5 & 0.0 & 0.7 & 6.2 \\
OpenClaw & 39.4 & 84.7\% & 173.0 & 91.0 & 5.0 & 0.5 & 2.7 \\
\bottomrule
\end{tabular*}
\end{table*}

\subsection{Trajectory and Efficiency Statistics}
\label{app:trajectory_cases}

Table~\ref{tab:trajectory_statistics} gives medians over the fixed 50-task set for all 18 configurations.
Runner-reported tokens are not fully comparable across harnesses because caching and subagent accounting differ.
Claude Code main trajectories also omit internal steps executed in isolated subagent contexts.
The table should therefore be read as a behavioral profile, not as a hardware-normalized efficiency leaderboard.

\begin{table*}[t]
\caption{Median process statistics over available task traces for all 18 configurations.
Steps count logged trajectory entries, and tool errors count failed invocations.
The \# columns are counts, time is in hours, and tokens are in millions.}
\label{tab:trajectory_statistics}
\centering
\scriptsize
\setlength{\tabcolsep}{3.6pt}
\begin{tabular*}{\columnwidth}{@{\extracolsep{\fill}}llrrrrr@{}}
\toprule
Harness & Model & Steps (\#) & Tool calls (\#) & Tool errors (\#) & Time (h) & Tokens (M) \\
\midrule
Claude Code & Claude Opus 5 & 260.5 & 148.5 & 5.5 & 3.9 & 28.7 \\
Claude Code & Claude Opus 4.8 & 286.0 & 131.0 & 2.0 & 1.9 & 21.5 \\
Claude Code & Claude Opus 4.7 & 221.0 & 149.0 & 2.0 & 2.1 & 15.1 \\
Claude Code & GPT-5.6 Sol & 167.0 & 114.5 & 11.5 & 1.3 & 4.6 \\
Claude Code & GPT-5.5 & 137.5 & 91.0 & 7.0 & 0.7 & 3.9 \\
Claude Code & Gemini-3.8-Flash & 237.0 & 125.0 & 9.0 & 3.1 & 10.9 \\
Claude Code & Qwen3.8-Omni-Flash & 373.5 & 183.5 & 5.0 & 5.1 & 14.8 \\
Claude Code & Qwen3.8-Max & 278.0 & 170.0 & 5.0 & 5.1 & 14.4 \\
Claude Code & Qwen3.7-Max & 156.0 & 105.5 & 7.0 & 2.4 & 4.3 \\
Claude Code & Qwen3.5-Omni-Plus & 151.5 & 84.5 & 14.0 & 0.4 & 4.4 \\
Claude Code & GLM-5.3 & 679.5 & 248.0 & 6.0 & 5.4 & 31.7 \\
Claude Code & Kimi-K3 & 248.0 & 142.0 & 3.0 & 4.6 & 11.0 \\
\midrule
Codex & GPT-5.6 Sol & 134.0 & 114.5 & 0.0 & 0.8 & 11.8 \\
Codex & GPT-5.5 & 179.0 & 122.0 & 0.0 & 0.6 & 6.2 \\
Codex & Qwen3.7-Max & 98.5 & 66.0 & 0.0 & 0.7 & 4.1 \\
\midrule
OpenClaw & GPT-5.6 Sol & 205.0 & 108.0 & 9.0 & 0.6 & 4.6 \\
OpenClaw & GPT-5.5 & 156.0 & 79.5 & 5.0 & 0.5 & 2.7 \\
OpenClaw & Qwen3.7-Max & 173.0 & 91.0 & 4.0 & 0.4 & 0.3 \\
\bottomrule
\end{tabular*}
\end{table*}

Table~\ref{tab:trajectory_order} reports ordering statistics for the same 18 configurations.
First write and last search are normalized event positions, where $0\%$ is the start of a trace and $100\%$ is the end.
Searches after write is the percentage of search calls that occur after the first report-writing event.
Write blocks count contiguous report-writing episodes separated by other actions.

\begin{table*}[t]
\caption{Median phase-order statistics over available task traces for all 18 configurations.
Positions and search shares are percentages.}
\label{tab:trajectory_order}
\centering
\scriptsize
\setlength{\tabcolsep}{3.6pt}
\begin{tabular}{llrrrrr}
\toprule
Harness & Model & Traces (\#) & First write pos. (\%) & Last search pos. (\%) & Post-write search (\%) & Write blocks (\#) \\
\midrule
Claude Code & Claude Opus 5 & 50 & 41.5 & 85.4 & 61.5 & 11.0 \\
Claude Code & Claude Opus 4.8 & 50 & 57.4 & 87.0 & 38.0 & 7.0 \\
Claude Code & Claude Opus 4.7 & 49 & 53.7 & 86.2 & 50.0 & 10.0 \\
Claude Code & GPT-5.6 Sol & 50 & 58.8 & 79.9 & 33.3 & 6.0 \\
Claude Code & GPT-5.5 & 50 & 60.8 & 79.4 & 25.0 & 4.5 \\
Claude Code & Gemini-3.8-Flash & 50 & 69.0 & 87.7 & 18.5 & 6.0 \\
Claude Code & Qwen3.8-Omni-Flash & 50 & 46.2 & 91.8 & 55.5 & 10.0 \\
Claude Code & Qwen3.8-Max & 49 & 46.9 & 88.1 & 50.0 & 9.0 \\
Claude Code & Qwen3.7-Max & 46 & 60.6 & 65.4 & 0.0 & 6.0 \\
Claude Code & Qwen3.5-Omni-Plus & 40 & 67.7 & 70.9 & 0.0 & 5.0 \\
Claude Code & GLM-5.3 & 50 & 50.1 & 92.3 & 54.2 & 15.5 \\
Claude Code & Kimi-K3 & 46 & 46.4 & 83.2 & 50.0 & 10.0 \\
\midrule
Codex & GPT-5.6 Sol & 50 & 75.6 & 90.5 & 14.3 & 6.0 \\
Codex & GPT-5.5 & 49 & 71.2 & 86.4 & 23.1 & 7.0 \\
Codex & Qwen3.7-Max & 50 & 62.2 & 80.0 & 21.4 & 5.0 \\
\midrule
OpenClaw & GPT-5.6 Sol & 43 & 60.6 & 78.8 & 34.8 & 2.0 \\
OpenClaw & GPT-5.5 & 48 & 23.9 & 79.9 & 100.0 & 3.5 \\
OpenClaw & Qwen3.7-Max & 41 & 69.1 & 88.3 & 16.7 & 4.0 \\
\bottomrule
\end{tabular}
\end{table*}

We label a configuration interleaved when its median share of searches after the first write is at least $30\%$ and its median write-block count is at least four.
Eight of 18 configurations meet this rule.
This threshold is used only to summarize a continuous pattern.
The underlying statistics are reported so that readers can apply another threshold.

The distributions show that a longer trace is neither necessary nor sufficient for a stronger report.
High-scoring systems tend to combine sustained evidence acquisition with successful transfer of that evidence into the report.
Low-scoring traces appear at both extremes, including short under-researched runs and long runs dominated by repeated repair or weak evidence integration.

\noindent\textbf{Long but weakly converted research.}
One Qwen3.8-Omni-Flash run uses 549 steps, 272 tool calls, 31.6 million runner-reported tokens, and 4.6 hours, yet scores 63.1.
The trace contains extensive video work, search, and report editing, but much of the acquired evidence does not become supported report claims.

\noindent\textbf{Organized delegated research.}
One GPT-5.6 Sol run under Claude Code uses 226 main-agent steps, 167 tool calls, and 7.9 million tokens and scores 71.5.
Its main trace uses explicit task checkpoints and 24 subagent calls for media reading, retrieval, and verification.
This case shows how delegation can keep the main context focused while retaining a long evidence path.

\noindent\textbf{Clean but shallow execution.}
One Qwen3.7-Max run under Codex uses 83 steps and 53 tool calls with no recorded tool error but scores 31.2.
The execution is stable, yet the limited research path leaves the report thin in evidence coverage and cross-source analysis.

\noindent\textbf{Repeated failure without repair.}
One Qwen3.5-Omni-Plus run uses 376 steps and 194 tool calls, records 126 tool errors, and scores 38.6.
Many errors repeat the same invalid validation path instead of changing the command or locating the correct file.
This loop consumes budget without improving the report.

\noindent\textbf{Fragmented but recoverable execution.}
One GPT-5.6 Sol run under OpenClaw uses 586 steps, 324 tool calls, and 16 tool errors and scores 72.1.
The trace decomposes retrieval and verification into many shell calls but still delivers a strong report.
This contrast shows why trajectory length alone cannot distinguish productive iteration from fragmentation.

These cases were selected after the aggregate analysis to illustrate recurring mechanisms.
They are not used as substitutes for system-level statistics.

\section{Prompt and Agent Protocols}
\label{app:prompts}

\definecolor{promptblue}{HTML}{2E7480}
\definecolor{promptpale}{HTML}{E8F2F5}
\definecolor{promptborder}{HTML}{A9D2D9}
\definecolor{promptink}{HTML}{253038}
\newcommand{\promptplate}[2]{%
  \par\medskip
  \begingroup
  \setlength{\fboxrule}{0.6pt}
  \setlength{\fboxsep}{7pt}
  \noindent\begin{minipage}{\linewidth}
  \noindent\colorbox{promptblue}{\parbox{\dimexpr\linewidth-2\fboxsep\relax}{\color{white}\bfseries #1}}
  \par\nointerlineskip
  \noindent\fcolorbox{promptborder}{promptpale}{\parbox{\dimexpr\linewidth-2\fboxsep-2\fboxrule\relax}{\small\color{promptink}\setlength{\parskip}{2pt}#2}}
  \end{minipage}
  \endgroup
  \par\medskip
}

This section records the stable prompt contracts used for task execution and benchmark construction.
Paths, model names, and instance metadata are filled automatically.
The released artifacts retain the complete prompts and machine-readable schemas.

\subsection{Evaluated Agent Instruction}

Every evaluated system receives the same deliverable request in the language of the source video.
The instruction identifies the local source video, its public source URL, the user focus, and the output directory.
It does not expose the evidence map, criterion atoms, or evaluator decisions.

\promptplate{Evaluated Agent Instruction}{
Use the video-research capability to conduct deep research on \texttt{[VIDEO]}.
\\
The public source is \texttt{[SOURCE URL]}.
\\
Focus on the following user need: \texttt{[QUERY]}.
\\
Produce an illustrated and source-cited HTML report in \texttt{[OUTPUT DIRECTORY]}.
\\
Retain detailed audiovisual notes, but make the search, analysis, and discussion answer the stated focus.
}

The high-level skill then supplies the seven-stage workflow described in Section~\ref{sec:workflow}.
Each ablation configuration receives only the capability groups listed in Table~\ref{tab:ablation_results}.
The fixed interface template is supplied separately and is not counted as a skill.

\subsection{Query Mining Prompts}

The query miner first constructs a timestamped audiovisual record.
Two isolated roles then receive that same record and the scenario label.
The video-view role is instructed as follows.

\promptplate{Video-View Query Mining Prompt}{
Propose five to ten research directions that originate from claims, theories, cases, procedures, parameters, or visual objects in this video.
\\
Each direction must require the given video, benefit from substantial external research, and remain broad enough for a report.
\\
Return the query, a video-grounded reason, and a confidence score.
}

The user-demand role receives a different contract.

\promptplate{User-Demand Query Mining Prompt}{
Propose five to ten realistic viewer needs involving a decision, diagnosis, reproduction, verification, or application problem.
\\
Each query must express the user's own task, depend on concrete audiovisual details in this video, and require external evidence beyond the video.
\\
Do not create a generic fact-check request by adding a superficial persona.
\\
Return the query, a video-grounded reason, and a confidence score.
}

The schema checker enforces separate arrays, five to ten candidates per view, a score in $[0,1]$, descending order, and language consistency.
Manual review then applies the boundary and persona tests described in Appendix~\ref{app:queries}.

\subsection{Independent Evidence Prompts}

The video-side collector is explicitly denied web access.
It is asked to transcribe the complete video, inspect frames and visible text, and emit one checkable claim per JSONL row with an absolute timestamp.
It must preserve the video's own wording and uncertainty instead of correcting claims from memory.

The web-side collectors receive only the query, domain, language, route description, and accumulated search seeds.
They cannot access the source video or another route's workspace.
Each collector searches one of six routes covering entities, provenance, concepts, people, analogues, and timelines.
It must open candidate sources, copy supporting text, mark source type and date, and preserve unavailable or conflicting evidence.

The common output contract is:
\promptplate{Atomic Evidence Record Contract}{
Emit one JSON object per atomic claim.
\\
Include \texttt{claim}, \texttt{type}, \texttt{origin}, \texttt{video\_ref}, \texttt{sources}, and \texttt{conflicts\_with}.
\\
Each source contains a canonical URL, source tier, publication date, and verbatim supporting quotation when available.
\\
Do not assign identifiers, confidence, difficulty, or relevance because later stages compute or adjudicate them.
}

\subsection{Rubric and Judge Prompts}

The atom generator receives the task specification and evidence records but no evaluated report.
It produces binary criteria that ask one checkable question and bind semantic criteria to fact identifiers.
The screening prompt rejects unsupported, compound, subjective, duplicate, ambiguous, or misassigned criteria.

At evaluation time, the judge receives one atom, its evidence note, and only the report material needed for that decision.
It must return Yes, No, or N/A with a short report-grounded rationale and quoted evidence when present.
It is instructed not to award credit from parametric knowledge or from evidence absent from the report.
J and M criteria use the same expected polarity and identifier scheme.
Model-judged gates use three independent votes.

\promptplate{Atomic Rubric Judge Contract}{
Given one binary criterion, its evidence note, and the relevant report material, decide whether the report satisfies the criterion.
\\
Return exactly one verdict from \texttt{Yes}, \texttt{No}, or \texttt{N/A}, followed by a short report-grounded rationale.
\\
Quote the supporting report span when it is present.
\\
Do not use parametric knowledge or evidence that does not appear in the submitted report.
\\
Preserve the supplied criterion identifier and expected polarity.
}

\section{Qualitative Task Examples}
\label{app:examples}

The examples below show how one source video determines the research boundary while external evidence changes the final value of the report.
They span decision support, technical explanation, scientific procedure, regulatory monitoring, and public-event analysis.
Table~\ref{tab:qualitative_tasks} summarizes the source-video evidence and external research required by each example.

\begin{table*}[t]
\caption{Representative tasks linking source-video and external evidence.}
\label{tab:qualitative_tasks}
\centering
\scriptsize
\setlength{\tabcolsep}{3.2pt}
\begin{tabular}{lp{0.27\textwidth}p{0.27\textwidth}p{0.32\textwidth}}
\toprule
Example & User focus & Video evidence & External research needed \\
\midrule
A & Decide whether a household should take additional mortgage leverage & The speaker's debt-to-asset claim and its implied measurement frame & Current mortgage rules, interest rates, household leverage data, price-to-income indicators, and decision uncertainty \\
B & Study the display and imaging trade-off of the shown under-display camera & Device identity, test conditions, and visible display or camera artifacts & Device specifications, optical mechanisms, independent tests, competing designs, and later product updates \\
C & Explain the pH jump and the distinction between stoichiometric and observed titration endpoints & The demonstrated curve, terminology, apparatus, and procedure & Textbook definitions, indicator ranges, measurement error, instrument guidance, and comparable demonstrations \\
D & Assess whether the shown renter-friendly plug-in pendant is safe and code compliant & Shown cord socket, PVC coupling, globe bulb, assembly steps, and visible product markings & Electrical code, listing standards, heat and strain risks, rental rules, safer alternatives, and current product evidence \\
E & Set alert and action limits for cleanroom viable colony-forming-unit (CFU) and particle data & The video's particle-count chart, its approximate thresholds, and the demonstrated monitoring context & Cleanroom standards, viable and nonviable monitoring guidance, historical process-control methods, and current regulatory requirements \\
\bottomrule
\end{tabular}
\end{table*}

\subsection{Report Organization}

\begin{figure*}[t]
  \centering
  \includegraphics[width=0.99\textwidth]{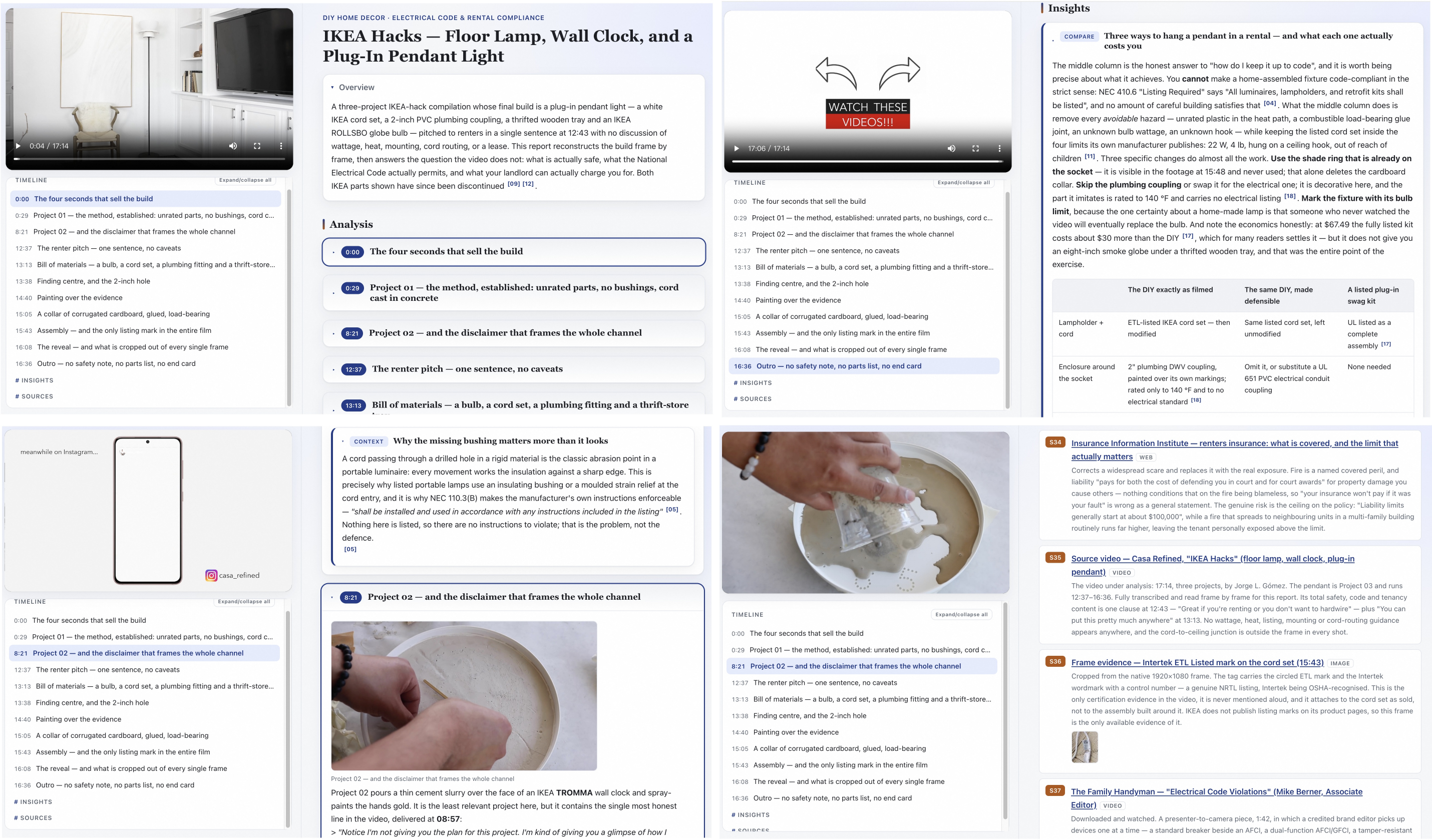}
  \caption{Four views of an interactive report for the pendant-light safety task.
  The interface aligns the audiovisual timeline with research findings, multimodal media, claim-level citations, and source records.}
  \label{fig:qualitative_report}
\end{figure*}

A valid report separates what the source video shows from what external evidence adds.
Each timeline section contains an absolute interval, an audiovisual note, and representative media.
Research findings cite external sources and state their relation to one or more video moments.
Comparisons and corrections retain uncertainty when sources disagree.
The interface lets a reader move from a note to the corresponding video time and from a claim to its cited source.
Figure~\ref{fig:qualitative_report} shows these elements in one generated report.

\subsection{Common Report Failures}

\noindent\textbf{Polished but incomplete.}
Some reports have strong hierarchy, imagery, and interaction, yet omit most evidence-map facts and comparisons.
This pattern motivates separate content and presentation dimensions.

\noindent\textbf{Source lists without claim support.}
Some reports contain many links, but the cited pages do not entail the nearby claims or are never connected to a source-video moment.
This pattern motivates separate citation count, closure, authority, and support directions.

\noindent\textbf{Transcript substitution.}
Some agents summarize speech and search its keywords without inspecting visible objects, charts, interfaces, motion, or other videos.
This pattern can produce a plausible report while bypassing the audiovisual evidence that defines the task.

\noindent\textbf{Research that never reaches the report.}
Long trajectories may open useful sources and inspect relevant media, but the final artifact can still omit the resulting facts or relations.
This pattern motivates evaluating evidence integration in the delivered report and analyzing process traces separately.

\section{Limitations and Future Work}
\label{app:limitations}

OmniDR-Bench currently covers 50 tasks across 19 subcategories, and future versions can collect more requests directly from users to expand task, scenario, and language coverage.
Some evaluation directions reflect capabilities emphasized by the reference skills, although scoring uses only the final report and Table~\ref{tab:ablation_results} compares several skill subsets.
Future work should develop more general and efficient skills, tools, and subagents that transfer across agent frameworks.
The current report interface uses a relatively fixed two-pane layout for audiovisual notes, research insights, and citations, which can be extended with richer structures and interactions.
Finally, this task can require hundreds of agent steps, so training and releasing capable open-source models remains important for the research community.

\end{document}